\documentclass[conference]{IEEEtran}

\IEEEoverridecommandlockouts                              

\usepackage[labelformat=simple]{subcaption}  
\usepackage{titletoc}
\usepackage[font=small]{caption}
\usepackage{graphicx} 
\usepackage{tabularx}
\usepackage{times} 
\usepackage{amsmath} 
\usepackage{amssymb}  
\usepackage{comment}
\usepackage{url}
\usepackage{xspace}
\usepackage{multirow}
\usepackage{colortbl}
\usepackage{booktabs}
\usepackage[table,xcdraw]{xcolor}
\usepackage[normalem]{ulem}
\newcolumntype{x}[1]{>{\centering\arraybackslash}p{#1pt}}
\newcolumntype{y}[1]{>{\raggedright\arraybackslash}p{#1pt}}
\newcolumntype{z}[1]{>{\raggedleft\arraybackslash}p{#1pt}}
\usepackage{xcolor}
\usepackage[table]{xcolor}  
\usepackage{pifont} 
\usepackage{array}
\usepackage{cite}
\usepackage{makecell}
\usepackage{enumerate}
\usepackage{tcolorbox}
\usepackage{listings}
\def\eg{\textit{e.g.}}

\makeatletter
\let\NAT@parse\undefined
\makeatother
\usepackage[pagebackref=false,breaklinks=true,letterpaper=true,colorlinks,bookmarks=false]{hyperref}
\usepackage[
  top=57pt,
  bottom=43pt,
  left=48pt,
  right=48pt,
  footskip=25pt,
]{geometry}
\newcommand{\tablestyle}[2]{\setlength{\tabcolsep}{#1}\renewcommand{\arraystretch}{#2}\centering\footnotesize}
\useunder{\uline}{\ul}{}
\useunder{\cellcolor[HTML]{EFEFEF}}{\hl}{}

\newcommand{\revise}[1]{#1}

\newcommand{\lookup}{\textsc{look-up}\xspace}
\newcommand{\lookdown}{\textsc{look-down}\xspace}

\title{\LARGE \bf
Stairway to Success: An Online Floor-Aware Zero-Shot Object-Goal \\ Navigation Framework via LLM-Driven Coarse-to-Fine Exploration 
}

\author{
    Zeying Gong$^1$~~\quad
    Rong Li$^1$~~\quad
    Tianshuai Hu$^2$~~\quad
    Ronghe Qiu$^1$~~\quad 
    Lingdong Kong$^3$
    \\
    Lingfeng Zhang$^4$~~\quad
    Guoyang Zhao$^1$~~\quad
    Yiyi Ding$^1$~~\quad 
    Junwei Liang$^{1,2,*}$ 
    \thanks{$^*$ Corresponding author. Email: {\tt\footnotesize junweiliang@hkust-gz.edu.cn}}
    \thanks{$^1$ The Hong Kong University of Science and Technology (Guangzhou)}
    \thanks{$^2$ The Hong Kong University of Science and Technology}
    \thanks{$^3$ National University of Singapore}
    \thanks{$^4$ Tsinghua University}
    \thanks{This work was supported by the Guangzhou Municipal Science and Technology Project under Grant 2024A03J0619.}
}

\begin{document}

\maketitle


\thispagestyle{empty}
\pagestyle{plain} 

\begin{abstract}
Deployable service and delivery robots struggle to navigate multi-floor buildings to reach object goals, as existing systems fail due to single-floor assumptions and requirements for offline, globally consistent maps. Multi-floor environments pose unique challenges including cross-floor transitions and vertical spatial reasoning, especially navigating unknown buildings. Object-Goal Navigation benchmarks like HM3D and MP3D also capture this multi-floor reality, yet current methods lack support for online, floor-aware navigation. To bridge this gap, we propose \textbf{\textit{ASCENT}}, an online framework for Zero-Shot Object-Goal Navigation that enables robots to operate without pre-built maps or retraining on new object categories. It introduces: (1) a \textbf{Multi-Floor Abstraction} module that dynamically constructs hierarchical representations with stair-aware obstacle mapping and cross-floor topology modeling, and (2) a \textbf{Coarse-to-Fine Reasoning} module that combines frontier ranking with LLM-driven contextual analysis for multi-floor navigation decisions. We evaluate on HM3D and MP3D benchmarks, outperforming state-of-the-art zero-shot approaches, and demonstrate real-world deployment on a quadruped robot. The project website is at \href{https://zeying-gong.github.io/projects/ascent/}{\textcolor{blue}{https://zeying-gong.github.io/projects/ascent/}}.
\end{abstract}


\section{INTRODUCTION}
\label{sec:intro}
Modern service and delivery robots are expected to operate in multi-floor buildings, from homes to offices. A simple command like \textit{``find the TV''} becomes challenging if the target is on another floor, requiring not only object recognition but also navigating across floors and reasoning vertically, particularly in unexplored buildings. This highlights a critical gap in robotics: \textbf{Object-Goal Navigation (OGN)} in multi-floor settings. However, existing online navigation systems often fail in these scenarios, limiting their real-world applications.

To quantify multi-floor navigation's importance, we analyze two widely-used OGN benchmarks, HM3D~\cite{ramakrishnan2021habitat} and MP3D~\cite{chang2017matterport3d}. Over half of validation episodes involve multi-floor buildings, with up to 28\% requiring explicit floor transitions (see Sec.~\ref{sec:exp_setting}). 
However, most OGN methods assume single-floor use. While multi-floor SLAM systems~\cite{chung2024nv} improve localization, they lack semantic reasoning for OGN in unseen scenes.

\begin{figure}[t]
    \centering
    \includegraphics[width=\linewidth]{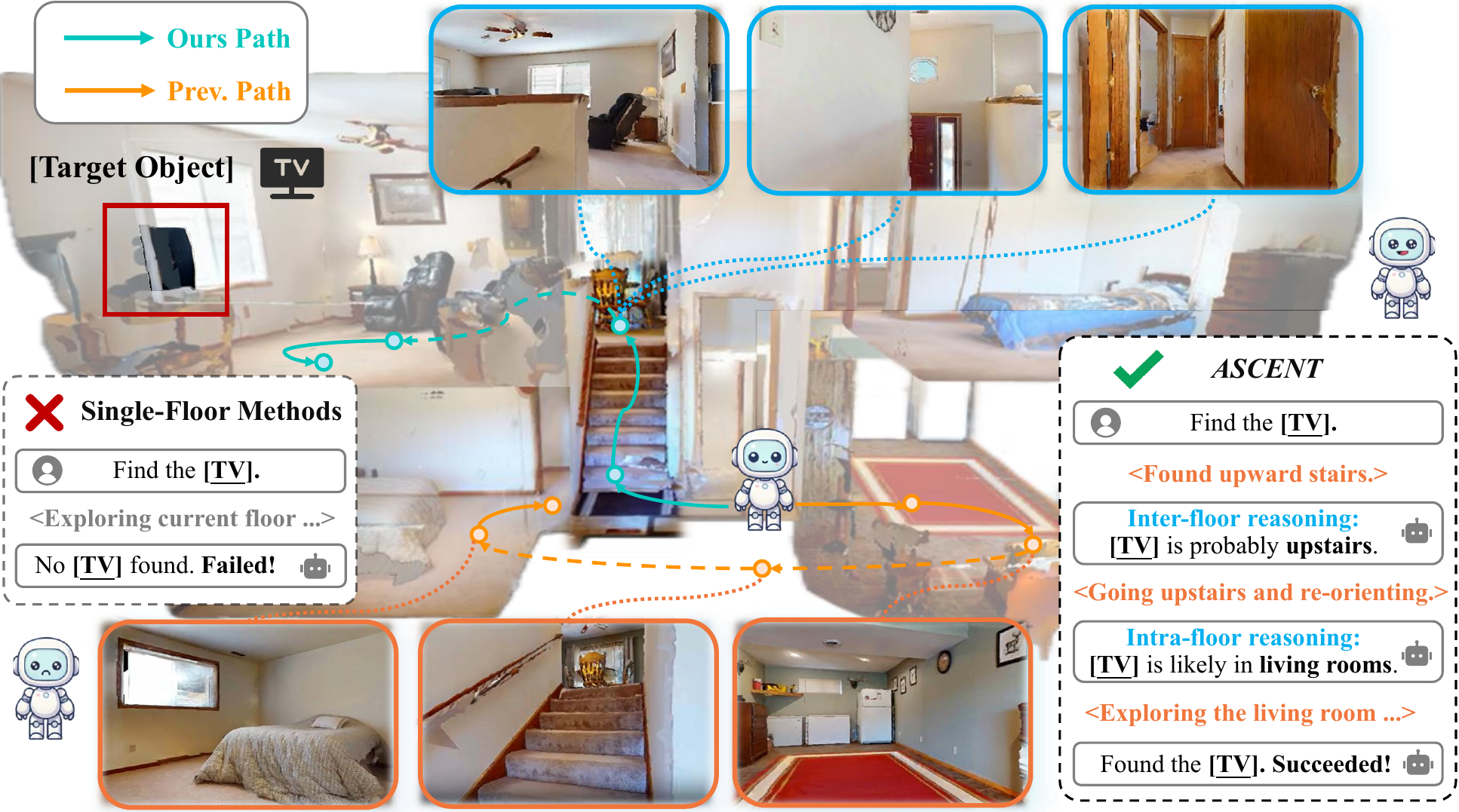}
    \caption{\textbf{Motivation of \textbf{\textit{ASCENT}.}} Unlike prior approaches that fail in multi-floor scenarios, our method enables online multi-floor navigation. By reasoning across floors, our policy succeeds in locating the goal and demonstrates a meaningful step forward in ZS-OGN.}
    \label{fig:teaser}
    \vspace{-0.2cm} 
\end{figure}

Moreover, recent OGN methods fall into two categories: (i) learning-based approaches~\cite{chen2023object,ramrakhya2023pirlnav,wasserman2024exploitation,yoo2024commonsense}, which achieve strong performance in familiar environments but require expensive data collection and retraining for unseen settings, and (ii) \textbf{Zero-Shot Object-Goal Navigation (ZS-OGN)} approaches~\cite{majumdar2022zson,guo2024object,yokoyama2024vlfm,yu2023l3mvn,yin2025sg,longinstructnav,cai2024bridging}. While ZS-OGN methods address the generalization limitation by operating without retraining, they still often neglect cross-floor planning and fail to handle multi-floor navigation effectively. Even recent multi-floor methods like MFNP~\cite{zhang2024multi} restrict exploration with one-way stair transitions, motivating our online multi-floor navigation method as shown in Fig. \ref{fig:teaser}.

Beyond multi-floor navigation, efficient planning poses another critical challenge for ZS-OGN. While Large Language Models (LLMs) have shown promise as high-level planners\cite{yu2023l3mvn,cai2024bridging,longinstructnav,yin2025sg}, frequent API calls lead to high costs and slow execution. An alternative is to use Vision-Language Models (VLMs) for local matching, as demonstrated by VLFM~\cite{yokoyama2024vlfm}. However, while its greedy strategy is faster, it lacks global planning, leading to oscillation and suboptimal paths.

To address these limitations, we propose \textbf{\textit{ASCENT}}, the first online framework that unifies dynamic multi-floor mapping with LLM-driven spatial reasoning for zero-shot navigation. Our approach presents a Multi-Floor Abstraction module that captures floor connectivity and enables inter-floor reasoning, allowing the robot to explore unseen buildings while maintaining semantic consistency across floors. By introducing a Coarse-to-Fine Reasoning module, we dramatically reduce the need for frequent LLM calls-by over 90\% compared to prior planners (see Sec. \ref{sec:experiment_abla}). This method leverages the VLM for coarse assessments while the LLM-based planner handles fine-grained decisions (\eg, floor/region selections), striking a balance between real-time performance and reasoning capability.

\begin{figure*}[ht]
    \centering
    \includegraphics[width=\linewidth]{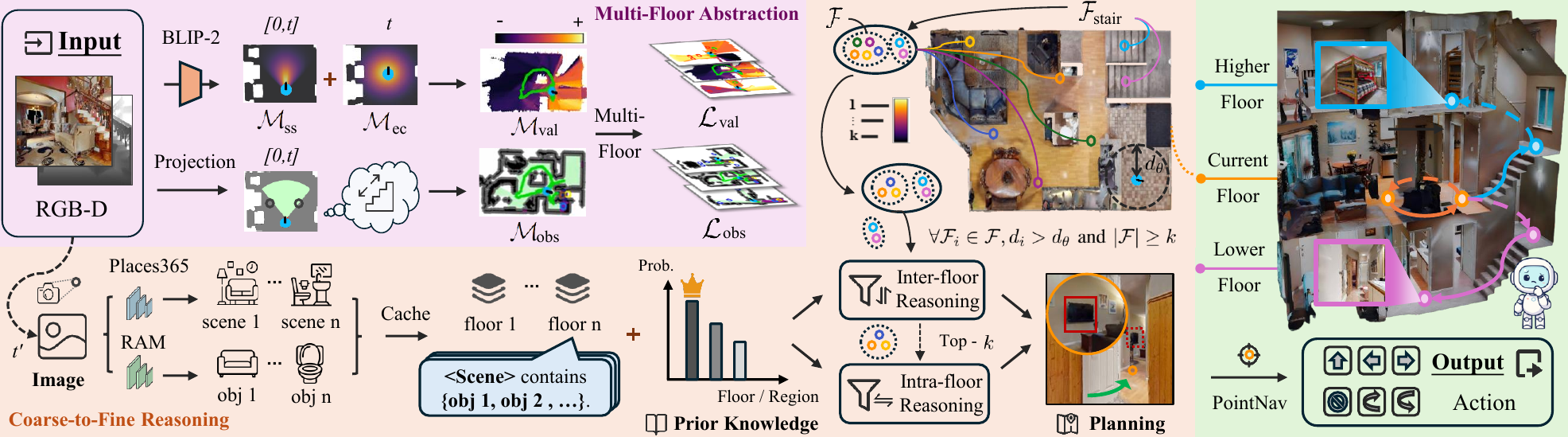}
    \caption{\textbf{Framework overview of \textbf{\textit{ASCENT}}}. The system takes RGB-D inputs (top-left), and outputs navigation actions (bottom-right). The Multi-Floor Abstraction module (top) builds intra-floor BEV maps and models inter-floor connectivity. The Coarse-to-Fine Reasoning module (bottom) uses the LLM for contextual reasoning across floors. Therefore, \textbf{\textit{ASCENT}} achieves floor-aware, Zero-Shot Object-Goal Navigation.}
    \label{fig:method_overview}
\vspace{-0.2cm}
\end{figure*}

To summarize, the key contributions of this work are:

\begin{itemize}
\item We introduce an online hierarchical framework for \textbf{multi-floor navigation without pre-built maps}, enabling exploration across floors in unseen environments.
\item We propose a coarse-to-fine frontier reasoning strategy that \textbf{reduces LLM calls by over 90\%} compared to prior works, while preserving strong planning performance.
\item We demonstrate \textbf{state-of-the-art (SOTA) performance} for ZS-OGN benchmarks, improving SR by 7.1\% and SPL by 6.8\% on HM3D, and SR by 3.4\% on MP3D. Beyond simulation, we validate the robustness through \textbf{real-world deployment} on a quadruped robot.
\end{itemize}
\section{RELATED WORK}
\subsection{Zero-Shot Object-Goal Navigation} 
ZS-OGN aims to find a target object in an unknown environment without any task-specific training. Early approaches, such as ZSON~\cite{majumdar2022zson}, leveraged VLMs like CLIP~\cite{radford2021learning} to transfer knowledge from Image-Goal to Object-Goal Navigation. \revise{Subsequent methods introduced modular designs: 
ActPept~\cite{guo2024object} jointly modeled geometric and semantic cues, 
VLFM~\cite{yokoyama2024vlfm} introduced vision-language frontier maps, and STRIVE~\cite{zhu2025strive} proposed structured representations with VLM-guided navigation. However, most existing methods assumed single-floor environments and lacked explicit multi-floor mechanisms.}

\subsection{\revise{LLMs for Object-Goal Navigation}} 
Recent ZS-OGN work has turned to LLMs for high-level planning. L3MVN~\cite{yu2023l3mvn} used LLMs to generate long-horizon plans, and SG-Nav~\cite{yin2025sg} constructed scene graphs for contextual memory. InstructNav~\cite{longinstructnav} proposed dynamic chain-of-navigation prompting, and PixNav~\cite{cai2024bridging} employed LLMs for room-level exploration. Beyond navigation planning, OpenFMNav~\cite{kuang2024openfmnav} and ApexNav~\cite{zhang2025apexnav} leverage LLMs for instruction parsing and detection enhancement respectively. Despite their strong reasoning, frequent LLM API calls cause high computational latency. \textbf{\textit{ASCENT}} addresses this limitation through Coarse-to-Fine Reasoning, significantly reducing LLM invocations while improving planning quality.

\subsection{Multi-Floor Navigation} 
Navigating across floors poses unique challenges, such as SLAM failures in repetitive or texture-poor environments~\cite{chung2024nv}. Early methods either assumed floor plans or relied on pre-built topological structures~\cite{kim2024development}. For example, Werby et al.~\cite{werby23hovsg} proposed hierarchical scene graphs for language-grounded navigation but required offline map construction. Recently, MFNP~\cite{zhang2024multi} pioneered multi-floor ZS-OGN but suffered from spatial map overlap, one-way stair constraints, and heuristic floor switching. We address these through per-floor representations, bidirectional traversal, and hierarchical planning.

\section{METHODOLOGY}
\label{sec:method}
This section introduces our proposed framework \textbf{\textit{ASCENT}} for the OGN task, with an overview shown in Fig. \ref{fig:method_overview}. Our approach formulates the OGN problem as minimizing a dual-cost objective that balances exploration and exploitation.
\begin{equation}
    \tau = \arg \min \Big( \underbrace{\lambda_{\mathrm{expl}} c_{\mathrm{expl}}(\tau)}_{\mathrm{exploration~cost}} + \underbrace{\lambda_{\mathrm{goal}} c_{\mathrm{goal}}(\tau)}_{\mathrm{exploitation~cost}} \Big)
\end{equation}

To achieve this dual objective, \textbf{\textit{ASCENT}} incorporates two key components. The \textbf{Multi-Floor Abstraction} module minimizes the exploitation cost $c_{\mathrm{goal}}$ by ensuring multi-floor accessibility and efficient goal-reaching (see Sec. \ref{subsec:multifloor_abstraction}). The \textbf{Coarse-to-Fine Reasoning} module reduces the exploration cost $c_{\mathrm{expl}}$ by prioritizing high-value frontiers (see Sec. \ref{subsec:c2f_frontier}).

\subsection{Multi-Floor Abstraction}
\label{subsec:multifloor_abstraction}

To support online navigation in multi-floor environments, our vision-based method eliminates the need for dedicated vertical sensors and abstracts complex 3D environments into a multi-layered representation to facilitate efficient planning. 
These core innovations through two key designs: \textbf{BEV Mapping Representations} and \textbf{Multi-Floor Topology Modeling}.

\subsubsection{\textbf{BEV Mapping Representations}}  
\label{subsubsec:singlefloor_map}  

Our approach builds on the concept of using Bird's Eye View (BEV) representations for frontier-based navigation, inspired by VLFM. Frontiers are defined as the midpoints of the boundaries between explored and unexplored areas, denoted as $\mathcal{F}$. For intra-floor exploration, we use two complementary representations:

\noindent\textbf{Exploration Value Map $\mathcal{M}_{\mathrm{val}}$:} This map guides efficient local exploration by integrating a \textit{Semantic Similarity Map} $\mathcal{M}_{\mathrm{ss}}$ with a proximity-based \textit{Exploration Cost Map} $\mathcal{M}_{\mathrm{ec}}$. \revise{Compared to VLFM, which only use $\mathcal{M}_{\mathrm{ss}}$, it ensures the robot fully exploits local, high-value areas before distant exploration.} The total value for the $i$-th frontier ${\mathcal{F}_i}$ is defined as:
\begin{equation}
\mathcal{M}_{\mathrm{val}}(\mathcal{F}_i) = \mathcal{M}_{\mathrm{ss}}(\mathcal{F}_i) + \begin{cases} \exp\left({- d_i }\right) & \text{if } d_i \le d_{\theta} \\ 0 & \text{otherwise} \end{cases}
\end{equation}

where $d_i$ represents the Euclidean distance from frontier $\mathcal{F}_i$ to the robot's position at the current moment $t$, $d_{\theta}$ is a distance threshold, and $\mathcal{M}_{\mathrm{ss}}$ aggregates values from $[0,t]$. 
\revise{The selected frontier is converted to a 2D waypoint goal and passed to the Point-Goal Navigation (PointNav) controller, which executes navigation for up to T/10 steps until the frontier is explored or the waypoint becomes unreachable.} Intuitively, this design encourages the robot to prioritize frontiers that are both semantically relevant (high $\mathcal{M}_{\mathrm{ss}}$) and spatially accessible (small $d_i$), ensuring that \revise{promising neighbor areas are exploited fully.}

\noindent\textbf{Stair-Aware Obstacle Map $\mathcal{M}_{\mathrm{obs}}$:} This is a binary occupancy grid constructed from depth data, which also aggregates values from $[0,t]$. 
\revise{Unlike VLFM, where obstacle maps treat stairs as impassable, ours re-labels stair-like structures as traversable after stair detection process. The map also maintains persistent blacklist/cache records of failed/successful transitions, providing the topological basis for multi-floor navigation. }

\subsubsection{\textbf{Multi-Floor Topology Modeling}}
\label{subsubsec:multifloor_model}

To enable online multi-floor navigation, our framework introduces two key designs: a \textbf{Stair Detection} module to identify traversable stairs and a \textbf{Cross-Floor Transition} module to execute inter-floor movements. Our approach provides superior robustness in both map representation and navigation policy compared to MFNP.

\begin{figure}[t]
    \centering
    \includegraphics[width=\linewidth]{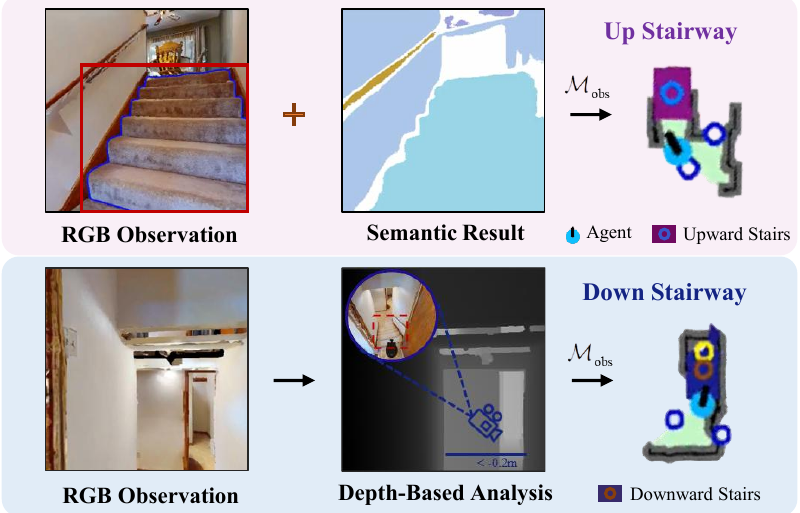}
    \caption{\textbf{Stair Detection process}. \textbf{\textit{ASCENT}} detects upward stairs (top) and infers downward stairs using depth-based analysis (bottom).}
    \label{fig:updown_stairs}
    \vspace{-0.2cm}
\end{figure}

\noindent\textbf{Stair Detection.}  
Our method detects traversable stairs by combining object detection with semantic segmentation. Let $S_{\mathrm{cand}}$ denote the set of staircase bounding boxes detected by the object detector, where each box $s$ has detection confidence $D(s)$. As shown in Fig. \ref{fig:updown_stairs}, a staircase candidate $s \in S_{\mathrm{cand}}$ is considered valid only when $D(s)$ exceeds threshold $\epsilon_1$ and the stair pixel proportion $\frac{\| \mathcal{P}_{\text{stair}}(s) \|}{\| \mathcal{P}_{\text{total}}(s) \|}$ exceeds $\epsilon_2$.
\begin{equation}
S_{\mathrm{valid}} = \left\{ s \in S_{\mathrm{cand}} \,\middle|\, D(s) > \epsilon_1 \land \frac{\| \mathcal{P}_{\text{stair}}(s) \|}{\| \mathcal{P}_{\text{total}}(s) \|} > \epsilon_\mathrm2 \right\}
\end{equation}

Furthermore, vision-based methods often fail to detect downward stairs in unexplored environments because their pixels are not visible from the current viewpoint, which is also a limitation of methods like MFNP. To address this, we identify areas with depth values less than -0.2m as potential downward stairs. The robot then executes a {LOOK DOWN} action and approaches the area to validate the potential stair, enabling robust bidirectional detection for seamless transitions.

\noindent\textbf{Cross-Floor Transition}. 
When a valid stair region is identified, the robot establishes a specific frontier $\mathcal{F}_{\mathrm{stair}}$ at its midpoint. When the robot makes a cross-floor decision, it first navigates toward $\mathcal{F}_{\mathrm{stair}}$ until reaching the stair area. Then its navigation target switches to a dynamic intermediate waypoint positioned at a distance of $d_{\mathrm{stair}}=0.8\,\text{m}$ ahead of its current pose. This encourages continuous forward movement while preventing premature stopping or localization drift. The low-level controller leverages a pre-trained PointNav policy.

For unexplored stairs, climbing continues until the robot exits the opposite side. For previously traversed stairs, the robot uses recorded start/end points for direct navigation. If climbing exceeds $\frac{T}{10}$ steps, the robot reverses to its previous position and marks the stair as impassable on $\mathcal{M}_{\mathrm{obs}}$. 

Upon floor transition, the robot updates its per-floor BEV map, forming map lists like $\mathcal{L}_{\mathrm{obs}} = [\mathcal{M}^{(1)}_{\mathrm{obs}}, ..., \mathcal{M}^{(N)}_{\mathrm{obs}}]$ and $\mathcal{L}_{\mathrm{val}} = [\mathcal{M}^{(1)}_{\mathrm{val}}, ..., \mathcal{M}^{(N)}_{\mathrm{val}}]$. Unlike MFNP that merges multi-floor data, our method prevents spatial overlap and inherently handles navigation across multiple floors. When multiple stairs connect the same floors, the robot explores new stairs initially but prioritizes previously traversed stairs or the largest stairs area for subsequent visits, ensuring robust floor revisits.

\revise{Overall, we integrate multiple feasibility checks: connectivity validation through $\mathcal{M}_{\mathrm{obs}}$, passability verification via dual-modality stair detection, and access control through blacklist/cache mechanisms, ensuring robust multi-floor navigation.}

\subsection{Coarse-to-Fine Reasoning}
\label{subsec:c2f_frontier}

To enable context-aware exploration in multi-floor environments, we introduce a two-stage reasoning pipeline: \textbf{Coarse-Grained Assessment} identifies high-value frontiers efficiently, and \textbf{Fine-Grained Decision} performs deeper contextual reasoning to select the final target. 
\revise{We leverage statistical priors from the training split of benchmarks to guide exploration, analogous to how learning-based methods capture in-domain knowledge without task-specific training.}

\subsubsection{\textbf{Coarse-Grained Assessment}} 

For computational efficiency, we first generate a set of distinct frontier proposals by evaluating image-text similarity within $\mathcal{M}_{\mathrm{val}}$. For each detected frontier, at its corresponding moment $t'$, the RGB image is processed to create a structured scene description using a scene classification model and an image tagging model. This description captures both the room type and associated objects, providing rich context. We then rank all available frontiers by their $\mathcal{M}_{\mathrm{val}}$ scores, and the top-$k$ proposals are cached for potential fine-grained analysis.

\subsubsection{\textbf{Fine-Grained Decision}}
\label{subsubsec:context_adapt}

\begin{figure}[t]
    \centering
    \includegraphics[width=\linewidth]{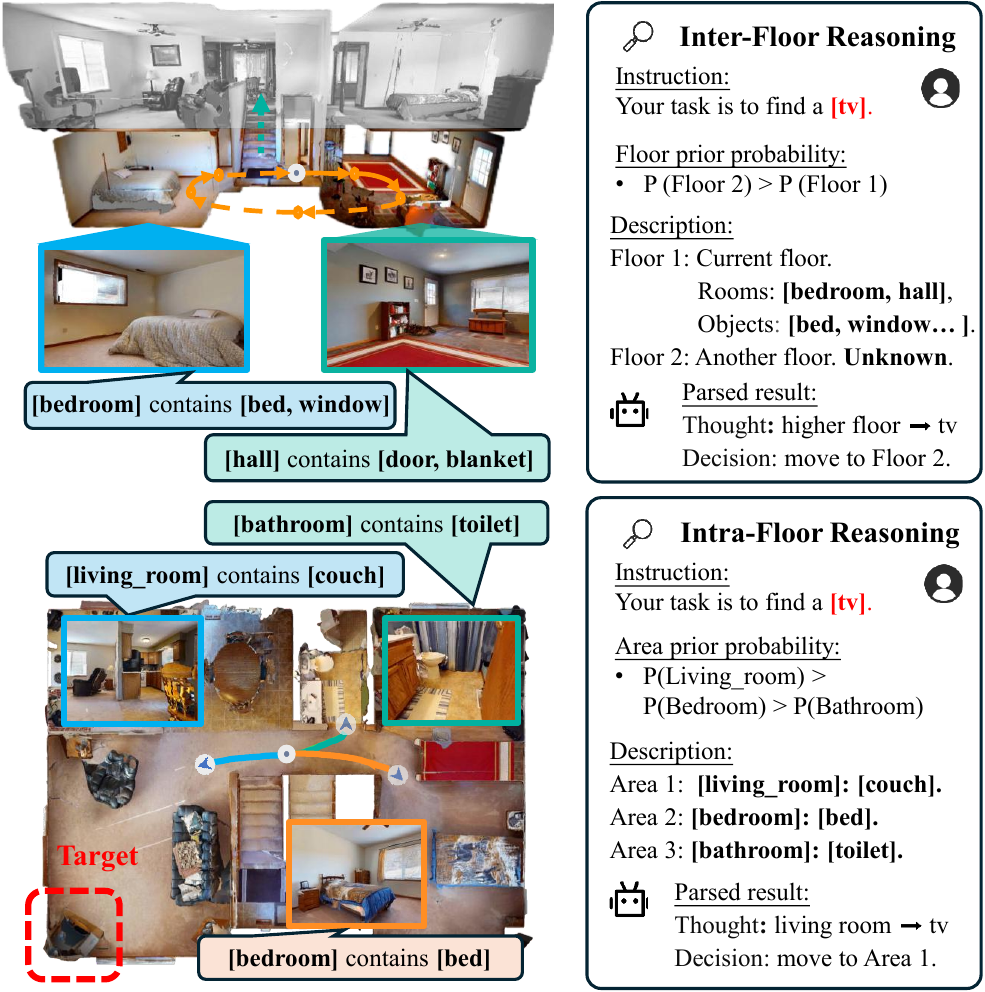}
    \caption{\textbf{Illustration of Fine-Grained Decision.} Following a coarse-grained assessment, the robot feeds cached contextual information and learned object priors to the LLM, which then decides whether to perform inter-floor transition or intra-floor navigation.} 
    \label{fig:fine_illu}
    \vspace{-0.2cm}
\end{figure}

By default, the robot directly selects the frontier with the highest score in $\mathcal{M}_{\mathrm{val}}$ for efficiency. This process is activated only when specific conditions are met: $\forall \mathcal{F}_i \in \mathcal{F}, d_i > d_\theta \text{ and } |\mathcal{F}| \geq k$, where $|\mathcal{F}|$ represents the total number of available frontiers and $k$ is the minimum frontier number threshold. Intuitively, this means LLM-based reasoning is triggered when no frontiers are nearby and sufficient frontier options exist for meaningful comparison. 
As shown in Fig. \ref{fig:fine_illu}, the robot performs contextual reasoning using cached information and LLMs in a sequential manner:

\noindent\textbf{Inter-Floor Reasoning}.
When stair-related frontiers $\mathcal{F}_{\mathrm{stair}}$ exist, the LLM first evaluates whether to switch floors by using contextual information from cached floor descriptions and learned object priors. A policy prevents frequent transitions by requiring an empirically-set minimum time interval $\frac{T}{10}$ steps or full floor coverage. If the decision is to switch floors, the robot proceeds with cross-floor transition.

\noindent\textbf{Intra-Floor Reasoning}. If the robot decides to remain on the current floor, the LLM analyzes semantic descriptions of available frontiers and prioritizes the most relevant location based on the task instruction (\eg, ``find the cabinet") and frontier context (\eg, room type, object presence).

This hierarchical design balances computational efficiency with contextual relevance, enabling multi-floor spatial reasoning while maintaining real-time performance.
\section{EXPERIMENTS}
\label{chapter:experiments}

In this section, we present a comprehensive experimental analysis to validate our \textbf{\textit{ASCENT}} framework. Our evaluation is designed to answer three key questions:

\begin{enumerate}
\item How well does it compare to existing methods? 
\item How does each module contribute to performance? 
\item Can it work successfully in real-world scenarios?
\end{enumerate}

\subsection{Experimental Settings}
\label{sec:exp_setting}
\begin{figure}[t]
    \centering
    \includegraphics[width=\linewidth]{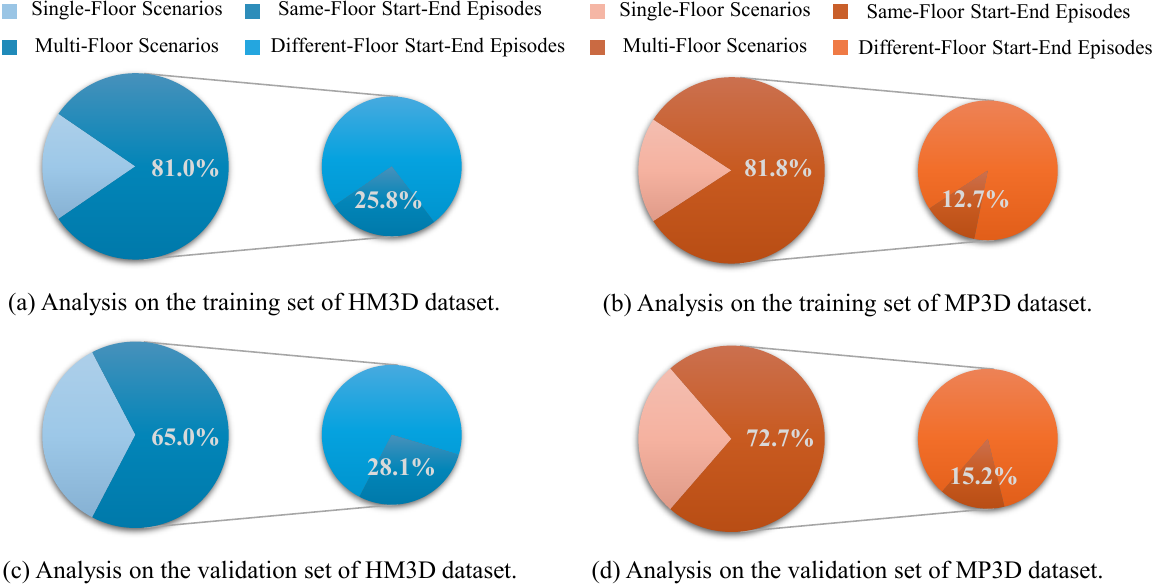}
    \caption{\textbf{Multi-floor scenario statistics in OGN benchmarks.} Across HM3D and MP3D, over half of scenarios involve multiple floors, with approximately one-fifth requiring cross-floor navigation.}
    \label{fig:multi_floor_stats}
    \vspace{-0.2cm}
\end{figure}

\noindent\textbf{Datasets.}
We evaluate our method on HM3D~\cite{ramakrishnan2021habitat} and MP3D~\cite{chang2017matterport3d}, the official Habitat Challenge benchmarks for OGN~\cite{habitatchallenge2022}. HM3D contains 2,000 validation episodes across 20 scenes with 6 object categories (all COCO classes), while MP3D includes 2,195 validation episodes across 11 scenes with 21 object categories (many non-COCO, open-vocabulary classes). As shown in Fig. \ref{fig:multi_floor_stats}, a significant portion of these environments are multi-floor: roughly 65\% of validation scenarios in HM3D and 73\% in MP3D. More importantly, a notable portion of episodes require cross-floor navigation, with targets located on different floors from the starting point in about 28\% of episodes in HM3D and 15\% in MP3D. These statistics establish the critical need for multi-floor navigation.

\noindent\textbf{Metrics.} 
We adopt two standard metrics from OGN evaluation~\cite{batra2020objectnav}: Success Rate (SR) and Success weighted by Path Length (SPL) to measure navigation performance and path efficiency. All results are reported in percentages.

\begin{table*}[t]
    \centering
    \caption{\revise{\textbf{Comparisons with SOTA methods.} 
The table contrasts learning-based and zero-shot methods on HM3D and MP3D datasets across the \textit{``Success Rate''} (SR) and \textit{``Success Weighted by Path Length''} (SPL) metrics. We introduce columns for \textit{Vision} and \textit{Language} models to specify the Instruction Interpolator components used by each approach. All baseline results from their original publications.}}
    \vspace{-0.1cm}
    \resizebox{0.85\linewidth}{!}{ 
    \tablestyle{4pt}{1.1}
    \begin{tabular}{x{35}|z{65}|z{60}|x{90}x{75}|cc|cc}
        \toprule
        & \multirow{2}{*}{\textbf{Method}} & \multirow{2}{*}{\textbf{Venue}} & \multicolumn{2}{c|}{\textbf{Instruction Interpolator}} & \multicolumn{2}{c|}{\textbf{HM3D} \cite{ramakrishnan2021habitat}} & \multicolumn{2}{c}{\textbf{MP3D} \cite{chang2017matterport3d}}
        \\
        & & & \textbf{Vision} & \textbf{Language}  & \textbf{SR} $\uparrow$ & \textbf{SPL} $\uparrow$ & \textbf{SR} $\uparrow$ & \textbf{SPL} $\uparrow$  
        \\
        \midrule\midrule
        \rowcolor{gray!10}\multicolumn{9}{c}{\textbf{\textcolor{gray}{Category: Learning-Based}}}
        \\
        \multirow{2}{*}{\begin{tabular}[r]{@{}c@{}} \textcolor{gray}{Single-} \\  \textcolor{gray}{Floor} \end{tabular}} 
        & \textcolor{gray}{RIM} \cite{chen2023object} & {\small \textcolor{gray}{IROS'23}} & \textcolor{gray}{-} & \textcolor{gray}{-} & \textcolor{gray}{$57.8$} & \textcolor{gray}{$27.2$} & \textcolor{gray}{$50.3$} & \textcolor{gray}{$17.0$} 
        \\
        & \textcolor{gray}{OVG-Nav} \cite{yoo2024commonsense} & {\small \textcolor{gray}{RAL'24}} & \textcolor{gray}{-} & \textcolor{gray}{-} & \textcolor{gray}{-} & \textcolor{gray}{-} & \textcolor{gray}{$35.8$} & \textcolor{gray}{$12.3$} 
        \\
        \midrule
        \multirow{2}{*}{\begin{tabular}[r]{@{}c@{}}  \textcolor{gray} {Multi-}\\  \textcolor{gray}{Floor} \end{tabular}}
        & \textcolor{gray}{PIRLNav} \cite{ramrakhya2023pirlnav} & {\small \textcolor{gray}{CVPR'23}} & \textcolor{gray}{-} & \textcolor{gray}{-} & \textcolor{gray}{$64.1$} & \textcolor{gray}{$27.1$} & \textcolor{gray}{-} & \textcolor{gray}{-} 
        \\
        & \textcolor{gray}{XGX} \cite{wasserman2024exploitation} & {\small \textcolor{gray}{ICRA'24}} & \textcolor{gray}{-} & \textcolor{gray}{-} & \textcolor{gray}{$72.9$} & \textcolor{gray}{$35.7$} & \textcolor{gray}{-} & \textcolor{gray}{-}
        \\
         \midrule\midrule
        \rowcolor{gray!10}\multicolumn{9}{c}{\textbf{Category: Zero-Shot}}
        \\
        \multirow{10}{*}{\begin{tabular}[r]{@{}c@{}} {Single-} \\ {Floor} \end{tabular}} &ZSON \cite{majumdar2022zson} & {\small NeurIPS'22} & CLIP \cite{radford2021learning} & - & $25.5$ & $12.6$ & $15.3$ & $4.8$
        \\
        & L3MVN \cite{yu2023l3mvn} & {\small IROS'23}  & - & \raisebox{-0.2\height}{\includegraphics[width=0.075\linewidth]{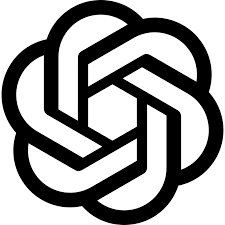}} GPT-2 \cite{liu2019roberta} & $50.4$ & $23.1$ & $34.9$ & $14.5$
        \\  
        & PixNav \cite{cai2024bridging} & {\small ICRA'24} & LLaMA-Adapter \cite{zhang2023llama} & \raisebox{-0.2\height}{\includegraphics[width=0.075\linewidth]{icons/gpt.png}} GPT-4 \cite{achiam2023gpt} & $37.9$ & $20.5$ & - & -
        \\
        & VLFM \cite{yokoyama2024vlfm} & {\small ICRA'24} & BLIP-2 \cite{li2023blip2} & - & $52.5$ & $30.4$ & $36.4$ & $17.5$
        \\
        & SG-Nav \cite{yin2025sg} & {\small NeurIPS'24} & LLaVA-1.6-7B \cite{liu2023visual} & \raisebox{-0.2\height}{\includegraphics[width=0.075\linewidth]{icons/gpt.png}} GPT-4 \cite{achiam2023gpt} & $54.0$ & $24.9$ & $40.2$ & $16.0$
        \\
        & \revise{OpenFMNav \cite{kuang2024openfmnav}} & \revise{{\small NAACL-F'24}} & \revise{\raisebox{-0.2\height}{\includegraphics[width=0.075\linewidth]{icons/gpt.png}} GPT-4V \cite{achiam2023gpt}} & \revise{\raisebox{-0.2\height}{\includegraphics[width=0.075\linewidth]{icons/gpt.png}} GPT-4 \cite{achiam2023gpt}} & \revise{$54.9$} & \revise{$24.4$} & \revise{$37.2$} & \revise{$15.7$}
        \\
        & ActPept \cite{guo2024object} & {\small RAL'24} & GraphSAGE \cite{hamilton2017inductive} & - & - & - & $39.8$ & $17.4$
        \\
        & InstructNav \cite{longinstructnav} & {\small CoRL'24} & \raisebox{-0.2\height}{\includegraphics[width=0.075\linewidth]{icons/gpt.png}} GPT-4V \cite{achiam2023gpt} & \raisebox{-0.2\height}{\includegraphics[width=0.075\linewidth]{icons/gpt.png}} GPT-4 \cite{achiam2023gpt} & $58.0$ & $20.9$ & - & - 
        \\
        & \revise{ApexNav \cite{zhang2025apexnav}} & \revise{{\small RAL'25}} & \revise{BLIP-2 \cite{li2023blip2}} & \revise{\raisebox{-0.2\height}{\includegraphics[width=0.11\linewidth]{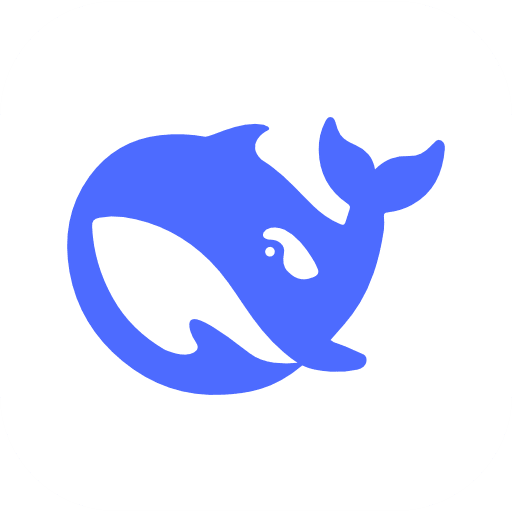}} DeepSeek-V3 \cite{liu2024deepseek}} & \revise{$59.6$} & \revise{$33.0$} & \revise{$39.2$} & \revise{$\mathbf{17.8}$}
        \\
        \midrule
        \multirow{2}{*}{\begin{tabular}[r]{@{}c@{}} {Multi-}\\ {Floor} \end{tabular}}&MFNP \cite{zhang2024multi} & {\small ICRA'25} & \raisebox{-0.2\height}{\includegraphics[width=0.06\linewidth]{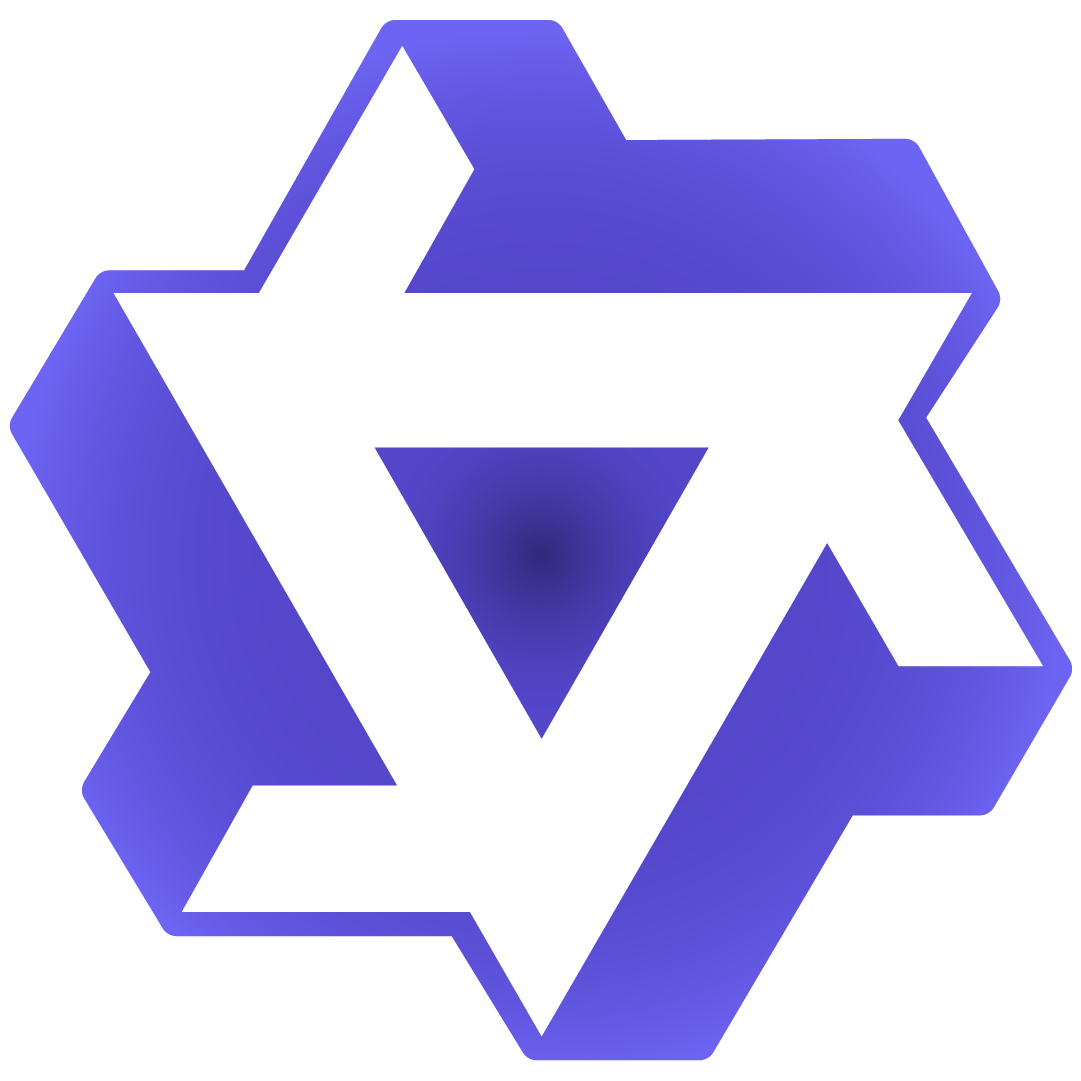}} Qwen-VLChat-Int4 \cite{bai2023qwen} & \raisebox{-0.2\height}{\includegraphics[width=0.085\linewidth]{icons/qwen.png}} Qwen2-7B \cite{team2024qwen2} & $58.3$ & $26.7$ & $41.1$ & $15.4$
        \\            
         & \cellcolor{gray!25}\textbf{\textit{ASCENT}} &  \cellcolor{gray!25}\textbf{Ours} &  \cellcolor{gray!25} BLIP-2 \cite{li2023blip2} &  \cellcolor{gray!25} \raisebox{-0.2\height}{\includegraphics[width=0.085\linewidth]{icons/qwen.png}} Qwen2.5-7B \cite{yang2024qwen2} &  \cellcolor{gray!25}  $\mathbf{65.4}$ &  \cellcolor{gray!25}  $\mathbf{33.5}$ &  \cellcolor{gray!25}  $\mathbf{44.5}$ &  \cellcolor{gray!25} ${15.5}$ \\
        \bottomrule
    \end{tabular}
    }
    \label{tab:main_table}
    \vspace{-0.2cm}
\end{table*}

\noindent\textbf{Implementation Details.}  
Our experiments are conducted in the Habitat~\cite{savva2019habitat}. \revise{For object detection, we use D-FINE~\cite{peng2024d} for COCO-class objects and Grounding-DINO~\cite{liu2024grounding} for open-set detection, following VLFM's confidence thresholds. Object segmentation uses Mobile-SAM~\cite{zhang2023faster}, and stair semantic segmentation uses RedNet~\cite{jiang2018rednet}.}
For frontier reasoning, we employ BLIP-2~\cite{li2023blip2} for semantic similarity, Places365~\cite{zhou2017places} for scene classification, RAM~\cite{zhang2024recognize} for object tagging, and Qwen2.5-7B-Instruct~\cite{yang2024qwen2} for LLM-based reasoning. 
\revise{Our method leverages two offline statistical priors: (1) Floor-level priors: distribution of goal objects across floor levels from training episodes of benchmarks. (2) Area-level priors: object-room co-occurrence probabilities from HM3D statistics. All experiments run on two NVIDIA RTX 3090 GPUs. }

\subsection{Main Results}

\noindent \textbf{Comparisons with SOTAs.} As shown in Tab. \ref{tab:main_table}, our method establishes a new SOTA in ZS-OGN on both HM3D and MP3D. 
\revise{The table reports all baseline results from their original publications and details the foundation models that serve as instruction interpolators,}
processing visual observations and language instructions for navigation decisions.

\revise{On HM3D, we achieve 65.4\% SR ($95\%$ CI: $[63.3\%, 67.5\%]$) and 33.5\% SPL ($95\%$ CI: $[31.3\%, 35.7\%]$), outperforming $58.3\%$ SR of MFNP ($95\%$ CI: $[56.2\%, 60.4\%]$) by +7.1\% SR and +6.8\% SPL. Bootstrap confidence intervals (10,000 iterations) confirm statistical significance with non-overlapping CIs. On MP3D, we reach 44.5\% SR ($95\%$ CI: $[42.4\%, 46.6\%]$) and 15.5\% SPL, surpassing $41.1\%$ SR of MFNP ($95\%$ CI: $[39.0\%, 43.2\%]$) by +3.4\% SR. Although CIs show minimal overlap, the consistent improvement across both datasets and controlled comparisons under identical model settings (see Tab.~\ref{tab:abla_large_model}) confirm that gains stem from architectural design. Compared to ApexNav, we trade path efficiency (-2.3\% SPL) for notably higher success (+5.3\% SR), as our cross-floor capability enables navigation in different-floor episodes where single-floor methods tend to fail.}

\noindent \textbf{Generalization Performance.} Supervised methods like RIM and XGX demonstrate high performance when trained on their corresponding datasets, achieving 50.3\% SR on MP3D and 72.9\% SR on HM3D, respectively. However, as shown in Tab. \ref{tab:generalization}, these methods exhibit limited cross-dataset generalization. In contrast, \textbf{\textit{ASCENT}} maintains consistent zero-shot performance across both datasets without any task-specific training. This strong cross-dataset consistency makes our framework well-suited for real-world deployment, where training data may not be available for unseen and complex environments.

\begin{table}[t]
\centering
\caption{\textbf{Generalization performance.} The SOTA learning-based methods show poor transferability, while our zero-shot method achieves strong cross-dataset generalization.}
\label{tab:generalization}
\begin{tabular}{l|c|cc|cc}
\toprule
\multirow{2}{*}{\textbf{Method}} 
& \multirow{2}{*}{\makecell{\textbf{Training Data}}} 
& \multicolumn{2}{c|}{\textbf{HM3D}}  
& \multicolumn{2}{c}{\textbf{MP3D}} \\
& & \textbf{SR} $\uparrow$ & \textbf{SPL} $\uparrow$ & \textbf{SR} $\uparrow$ & \textbf{SPL} $\uparrow$ \\
\midrule
RIM~\cite{chen2023object} & MP3D & $57.8$ & $27.2$ & $\mathbf{50.3}$ & $\mathbf{17.0}$ \\
XGX~\cite{wasserman2024exploitation} & HM3D & $\mathbf{72.9}$ & $\mathbf{35.7}$ & $13.6$ & $5.1$ \\
\midrule
\rowcolor{gray!25}\textbf{\textit{ASCENT}} & -- & $65.4$ & $33.5$ & $44.5$ & $15.5$ \\
\bottomrule
\end{tabular}
\vspace{-0.1cm}
\end{table}

\noindent\textbf{Performance Analysis Across Floor Scenarios.}
As shown in Tab. \ref{tab:multi-floor}, \textbf{\textit{ASCENT}} demonstrates distinct advantages across different floor scenarios. For same-floor start-end episodes, our method outperforms both baselines, validating the effectiveness of our Coarse-to-Fine Reasoning. The advantages are even more pronounced in different-floor start-end episodes, where target objects are located on a different floor from the robot's starting point. \textbf{\textit{ASCENT}} achieves 33.3\% SR, substantially outperforming VLFM (+32.9\% SR) and significantly surpassing MFNP (+19.9\% SR). This substantial improvement validates our framework's effectiveness in handling complex multi-floor transitions. Additionally, \textit{ASCENT-Ideal} achieves 49.8\% SR in inter-floor episodes, demonstrating that nearly half of these challenging cross-floor navigation tasks could be resolved with improved object detection, highlighting our algorithm's potential beyond current perception constraints.

\begin{table}[t]
    \centering
    \scriptsize 
    \caption{\textbf{Performance comparison across floor scenarios.} Start-End refers to robot initial position and target location. All conducted on HM3D, and results marked with \textdagger\ from author correspondence.}
    \resizebox{\linewidth}{!}{
    \begin{tabular}{l|cc|cc|cc}
        \toprule
        \multirow{2}{*}{\textbf{Method}} & \multicolumn{2}{c|}{\textbf{Same-Floor Start-End}} & \multicolumn{2}{c|}{\textbf{Different-Floor Start-End}} & \multicolumn{2}{c}{\textbf{All Episodes}} \\
         & \textbf{SR}$\uparrow$   & \textbf{SPL}$\uparrow$ & \textbf{SR}$\uparrow$   & \textbf{SPL}$\uparrow$ & \textbf{SR}$\uparrow$   & \textbf{SPL}$\uparrow$\\
        \midrule
        VLFM & 64.6 & 37.3 & 0.4 & 0.1 & 52.8 & 30.5 \\
        MFNP\textdagger & $68.4$ & $30.5$ & $13.4$ & $9.8$  & $58.3$ & $26.7$ \\
        \midrule
        \rowcolor{gray!25}\textbf{\textit{ASCENT}} & $72.6$ & $37.7$ & $33.3$ & $14.9$ & $65.4$ & $33.5$ \\
        \textit{ASCENT-Ideal} & $\mathbf{74.9}$ & $\mathbf{45.1}$ & $\mathbf{49.8}$ & $\mathbf{22.6}$ & $\mathbf{70.3}$ & $\mathbf{41.0}$ \\
        \bottomrule
    \end{tabular}}
    \label{tab:multi-floor}
    \vspace{-0.2cm}
\end{table}

\subsection{Component Analysis}
\label{sec:experiment_abla}

\noindent \textbf{Ablation Study.} To evaluate the contribution of each core component, we conducted a detailed ablation study on the full \textbf{\textit{ASCENT}} framework. Tab. \ref{tab:ablation} presents the results on the HM3D dataset, which clearly demonstrates that all components are crucial for achieving optimal performance. The Exploration Cost Map has the most significant impact, its removal leads to a substantial performance loss of $9.1\%$ in SR and $5.8\%$ in SPL, as the robot inefficiently traverses frontiers. Similarly, removing the Cross-Floor Transition module impairs the robot's ability to navigate multi-floor environments, resulting in a drop of $8.7\%$ in SR and $4.9\%$ in SPL. The absence of Coarse-to-Fine Reasoning causes a loss of $7.7\%$ in SR and $5.0\%$ in SPL, as the robot becomes trapped in local optima by relying solely on instantaneous semantic matching.
Finally, although the variant without Prior Knowledge still outperforms the SOTA method MFNP (+3.8\% SR), incorporating statistical priors on object distribution further refines the LLM's decision-making, yielding an additional $3.3\%$ increase in SR. Overall, the integration of all four modules allows \textbf{\textit{ASCENT}} to achieve improvements of $12.9\%$ and $3.1\%$ over the baseline, confirming the necessity and effectiveness of each component.

\begin{table}[t]
    \centering
    \caption{\revise{\textbf{Ablation study on core components.} Each row removes one core component from the full \textbf{\textit{ASCENT}} model.}}
    \label{tab:ablation}
    \resizebox{\linewidth}{!}{
    \tablestyle{9pt}{1.1}
    \begin{tabular}{lcccc}
    \toprule
    \multirow{2}{*}{\textbf{Method}} & \multicolumn{2}{c}{\textbf{HM3D}} & \multicolumn{2}{c}{\textbf{MP3D}} \\ 
\cmidrule(lr){2-3} \cmidrule(lr){4-5} 
    & \textbf{SR} $\uparrow$ & \textbf{SPL} $\uparrow$ & \textbf{SR} $\uparrow$ & \textbf{SPL} $\uparrow$ \\ 
\midrule 
    w/o Exploration Cost Map & $56.3$ & $27.7$ & $37.6$ & $13.1$
    \\
    w/o Cross-Floor Transition & $56.7$ & $28.6$ & $39.9$ & $\mathbf{17.9}$
    \\
    w/o Coarse-to-Fine Reasoning & $57.7$ & $28.5$ & $40.8$ & $14.2$
    \\  
    w/o Prior Knowledge & $62.1$ & $30.1$ & $42.5$ & $15.5$ 
    \\ 
    w/o Floor-Level Prior Knowledge & $62.7$ & $31.4$ & $43.9$ & $16.3$ 
    \\
    w/o Area-Level Prior Knowledge & $63.8$ & $32.9$ & $43.3$ & $15.0$ 
    \\ 
    \midrule
    \rowcolor{gray!25}\textbf{\textit{ASCENT}} & $\mathbf{65.4}$ & $\mathbf{33.5}$ & $\mathbf{44.5}$ & $15.5$
    \\ 
    \bottomrule
    \end{tabular}}
    \vspace{-0.1cm}
\end{table}

Beyond the core ablation study, we provide a deeper quantitative analysis of our method's performance across different large model configurations and object detector choices.

\begin{table}[t]
\centering
\caption{\revise{\textbf{Effect of large model configurations.} Results marked with \textdagger\ from author correspondence.}}
\label{tab:abla_large_model}
\resizebox{\linewidth}{!}{
\begin{tabular}{l|cc|cc|cc}
\toprule
\multirow{2}{*}{\textbf{Method}} & \multicolumn{2}{c|}{\textbf{Instruction Interpolator}} & \multicolumn{2}{c|}{\textbf{HM3D}} & \multicolumn{2}{c}{\textbf{MP3D}} \\
& \textbf{Vision} & \textbf{Language} & \textbf{SR}$\uparrow$ & \textbf{SPL}$\uparrow$ & \textbf{SR}$\uparrow$ & \textbf{SPL}$\uparrow$ \\
\midrule
MFNP\textdagger & BLIP-2 & Qwen2.5-7B & 55.8 & 24.9 & 38.5 & 12.4 \\
MFNP & \cellcolor{gray!25}Qwen-VLChat-Int4 & \cellcolor{gray!25}Qwen2-7B & \cellcolor{gray!25}58.3 & \cellcolor{gray!25}26.7 & \cellcolor{gray!25}41.1 & \cellcolor{gray!25}15.4 \\
\midrule
\multirow{4}{*}{\textit{ASCENT}}
 & BLIP-2 & Qwen2-7B & $65.2$ & $33.3$ & $44.1$ & $15.2$ \\
& \cellcolor{gray!25}BLIP-2 & \cellcolor{gray!25}Qwen2.5-7B & \cellcolor{gray!25}$65.4$ & \cellcolor{gray!25}$33.5$ & \cellcolor{gray!25}$44.5$ & \cellcolor{gray!25}$15.5$ \\
 & Qwen-VLChat-Int4 & Qwen2-7B & $67.7$ & $34.9$ & $46.5$ & $18.6$ \\
 & Qwen-VLChat-Int4 & Qwen2.5-7B & $\mathbf{67.9}$ & $\mathbf{35.0}$ & $\mathbf{46.8}$ & $\mathbf{18.8}$ \\
\bottomrule
\end{tabular}}
\vspace{-0.2cm}
\end{table}

\noindent\textbf{Effect of Large Model Components.} 
As shown in Tab. \ref{tab:abla_large_model}, we evaluate how different large model components affect navigation performance. Replacing MFNP’s original models (Qwen-VLChat-Int4 + Qwen2-7B) with our configuration (BLIP-2 + Qwen2.5-7B), surprisingly results in a performance drop from 58.3\% to 55.8\% SR on HM3D. This suggests that foundation model choice alone is insufficient and requires proper architectural integration. Conversely, our method proves robust across different model configurations. \textbf{\textit{ASCENT}} achieves 65.4\% SR, while \textit{ASCENT}* with the original MFNP components further improves to 67.7\% SR and 34.9\% SPL. This confirms that our gains stem primarily from architectural innovation, not merely stronger foundation models.

\noindent\textbf{Effect of Object Detectors.} 
As shown in Tab.~\ref{tab:abla_obj_detector}, \textbf{\textit{ASCENT}}'s performance scales with detector quality, revealing its theoretical upper bound under ideal perception: it achieves up to 70.3\% SR on HM3D and 58.6\% SR on MP3D. The performance gap on MP3D (+16.2\% SR) is larger than that on HM3D (+9.4\% SR), due to broader open-vocabulary challenges on MP3D. Importantly, under identical detection settings (e.g., G-DINO + YOLOv7), \textbf{\textit{ASCENT}} consistently outperforms VLFM (by +8.4\% SR on HM3D and +6.0\% SR on MP3D), demonstrating that its gains stem from architectural design rather than perception advantages. 

\begin{table}[t] 
\centering 
\caption{\revise{\textbf{Effect of object detectors.} Better detectors lead to better performance. G-DINO represents GroundingDINO detector.}} 
\begin{tabular}{lccccc} 
\toprule 
\multirow{2}{*}{\textbf{Method}} & \multirow{2}{*}{\textbf{Object Detector}} & \multicolumn{2}{c}{\textbf{HM3D}} & \multicolumn{2}{c}{\textbf{MP3D}} \\ 
\cmidrule(lr){3-4} \cmidrule(lr){5-6} 
& & \textbf{SR} $\uparrow$ & \textbf{SPL} $\uparrow$ & \textbf{SR} $\uparrow$ & \textbf{SPL} $\uparrow$ \\ 
\midrule 
\multirow{3}{*}{VLFM} & G-DINO + YOLOv7 & 52.5 & 30.4 & 36.4 & 17.5 \\ 
& G-DINO + D-FINE & 54.1 & 33.0 & 37.5 & 17.8 \\ 
& Ideal & 62.4 & 40.0 & 55.3 & 29.6 \\ 
\midrule 
\multirow{3}{*}{\textbf{\textit{ASCENT}}} 
& G-DINO + YOLOv7 & 60.9 & 29.6 & 42.4 & 13.8 \\ 
& \cellcolor{gray!25}G-DINO + D-FINE & \cellcolor{gray!25}65.4 & \cellcolor{gray!25}33.5 & \cellcolor{gray!25}44.5 & \cellcolor{gray!25}15.5 \\ 
& Ideal & \textbf{70.3} & \textbf{41.0} & \textbf{58.6} & \textbf{30.3} \\ 
\bottomrule 
\end{tabular}
\label{tab:abla_obj_detector} 
\vspace{-0.1cm} 
\end{table}

\subsection{Efficiency \& Sensitivity Analysis}
\noindent\textbf{Computational Efficiency Comparison.}
To evaluate the practical deployment feasibility of our method, we compare computational efficiency with recent LLM-based planners on randomly selected HM3D scenarios (3 single-floor and 3 multi-floor scenarios). Note that MFNP is not open-source and thus excluded from this comparison. As shown in Tab. \ref{tab:efficiency}, \textbf{\textit{ASCENT}} demonstrates superior efficiency across multiple metrics. Our method reduces LLM calls by over 90\% compared to existing LLM planners (2.0-2.7 vs. 35.0-149.4 calls per episode), while achieving higher SR, better SPL and shorter navigation times. This efficiency advantage stems from our strategic LLM usage design, where language models are invoked only for high-level reasoning rather than step-by-step planning. The reduced computational overhead enables real-time deployment while maintaining navigation performance.

\begin{table}[t]
\centering
\caption{\textbf{Computational efficiency comparison.} Results are averaged over randomly selected scenarios on HM3D. *Methods reimplemented with Qwen2.5-32B due to GPT-4 API deprecation.}
\label{tab:efficiency}
\resizebox{\linewidth}{!}{
\begin{tabular}{l|cccc|cccc}
\toprule
\multirow{2}{*}{\textbf{Method}} & \multicolumn{4}{c|}{\textbf{Single-Floor Scenario}} & \multicolumn{4}{c}{\textbf{Multi-Floor Scenario}} \\
& \textbf{SR}$\uparrow$ & \textbf{LLM Calls}$\downarrow$ & \textbf{Steps}$\downarrow$ & \textbf{Runtime (s)}$\downarrow$ & \textbf{SR}$\uparrow$ & \textbf{LLM Calls}$\downarrow$ & \textbf{Steps}$\downarrow$ & \textbf{Runtime (s)}$\downarrow$ \\
\midrule
L3MVN & 51.5 & 35.0 & 194.1 & 1047.7 & 48.5 & 38.8 & 216.5 & 1173.1 \\
PixNav* & 48.2 & 58.1 & 278.7 & 276.4 & 48.1 & 62.8 & 287.5 & 290.1 \\
SG-Nav & 54.2 & 122.3 & 300.0 & 728.5 & 46.5 & 149.4 & 250.8 & 323.6 \\
InstructNav* & 54.4 & 113.6 & 320.5 & 458.8 & 54.3 & 108.3 & 446.5 & 450.1 \\
\midrule
\rowcolor{gray!25}\textbf{\textit{ASCENT}} & $\mathbf{57.6}$ & $\mathbf{2.0}$ & $\mathbf{171.4}$ & $\mathbf{190.3}$ & $\mathbf{64.8}$ & $\mathbf{2.7}$ & $\mathbf{153.3}$ & $\mathbf{181.6}$ \\
\bottomrule
\end{tabular}}
\vspace{-0.2cm}
\end{table}

\begin{figure}[ht]
    \centering
    \includegraphics[width=\linewidth]{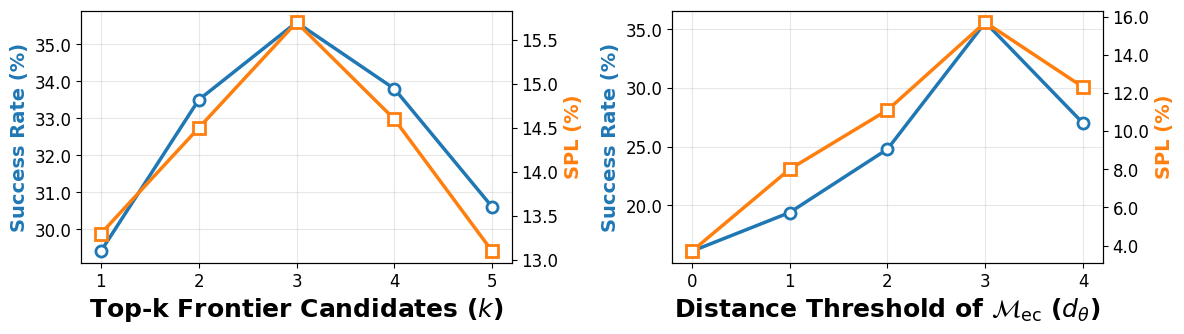}
    \caption{\textbf{Hyperparameter sensitivity analysis.} Grid search on three scenes of MP3D: optimal performance at $k=3$ and $d_{\theta}=3.0$.}
\label{fig:hyperparameter}
\vspace{-0.2cm}
\end{figure}

\noindent\textbf{Hyperparameter Sensitivity Analysis.}
We conduct a systematic sensitivity analysis for key hyperparameters in our framework. As shown in Fig. \ref{fig:hyperparameter}, we evaluate the impact of top-k frontier candidates ($k$) and distance threshold of Exploration Cost Map $\mathcal{M}_{\mathrm{ec}}$ ($d_\theta$) through grid search on three MP3D validation scenes. The parameter $k$ controls the number of frontier candidates considered by the LLM in Coarse-to-Fine Reasoning, while $d_\theta$ determines the spatial range in $\mathcal{M}_{\mathrm{val}}$ for Multi-Floor Abstraction. The results reveal that $k=3$ and $d_\theta=3.0$ yield optimal performance. Therefore, we adopt this combination of hyperparameters for all other experiments.

\subsection{Real-World Deployment}
\label{sec:realworld}

\begin{figure*}[t]
    \centering
    \includegraphics[width=\linewidth]{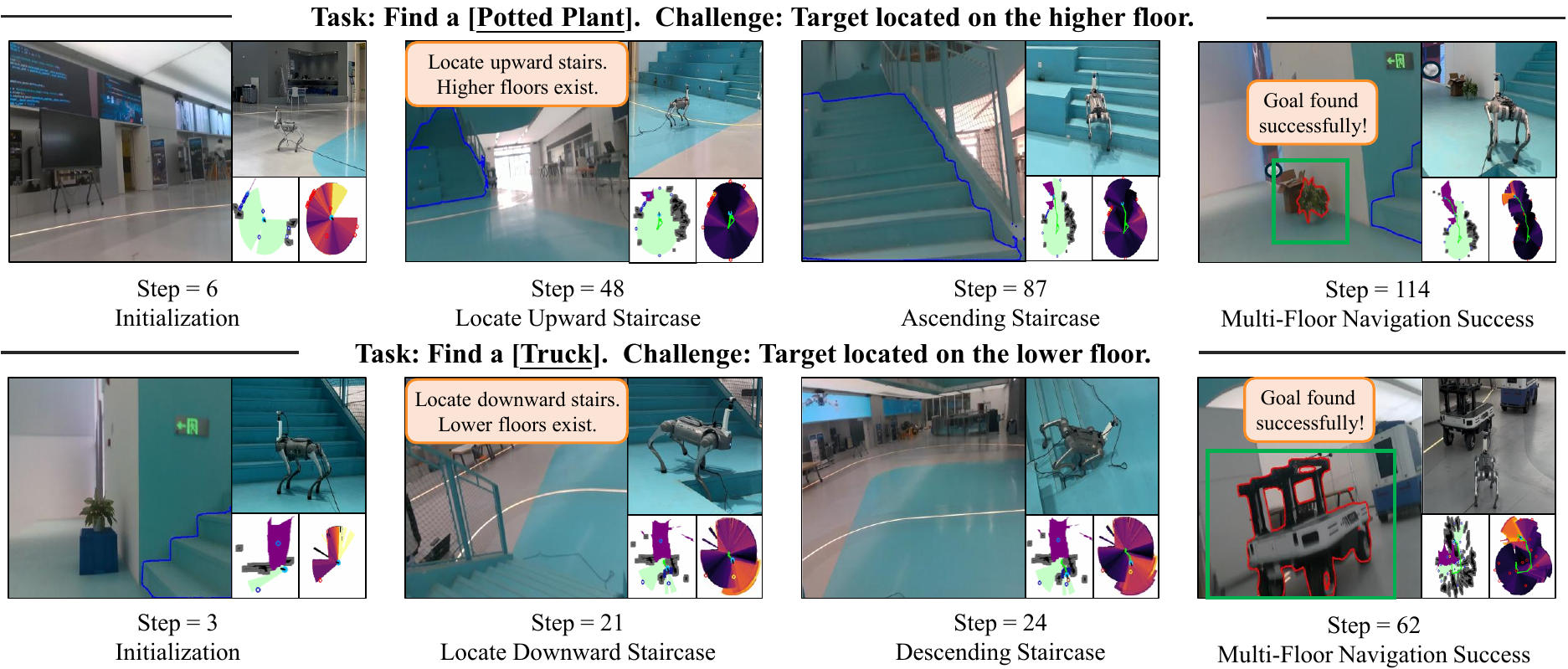}
    \caption{\textbf{Real-world deployment.} We conduct real-world experiments for \textbf{\textit{ASCENT}} method on the Unitree quadruped robot \textit{Go2}. The top row shows a successful multi-floor navigation task where the robot starts on the first floor and ascends stairs to find a ``potted plant". The bottom row illustrates a successful descent task to find a ``truck" on a lower floor. The figure highlights key steps of the mission, including stairway localization, cross-floor navigation, and target detection, demonstrating the framework's ability to handle complex multi-floor scenarios.}
\label{fig:reality_demo}
\vspace{-0.2cm}
\end{figure*}

We deploy \textbf{\textit{ASCENT}} on a Unitree Go2 quadruped robot to validate its effectiveness in real-world multi-floor scenarios. As shown in Fig.~\ref{fig:reality_demo}, the robot successfully navigates multi-floor environments, ascending stairs to find a ``potted plant" and descending to locate a ``truck", validating our method's generalization from simulation to reality.

\begin{table}[ht]
\centering
\caption{\textbf{Real-world experimental results.} Values
are averaged over all trials. TD: Traveled Distance (m); TT: Traveled Time (s).}
\label{tab:reality_stat}
{
\begin{tabular}{l|cccc}
\toprule
\textbf{Scenario} & \textbf{TD} & \textbf{TT} & \textbf{SR}$\uparrow$ & \textbf{SPL}$\uparrow$ \\
\midrule
Current-Floor & 8.3 & 75.7 &  62.5 & 35.9 \\
Higher-Floor & 16.9 & 200.8 & 37.5 & 20.2 \\
Lower-Floor & 20.5 & 264.1 & 25.0 & 12.4 \\
\midrule
\textbf{Total} & 15.2  & 180.2 & 41.7 & 22.8 \\
\bottomrule
\end{tabular}}
\vspace{-0.2cm}
\end{table}
\begin{figure}[ht]
    \centering
    \includegraphics[width=\linewidth]{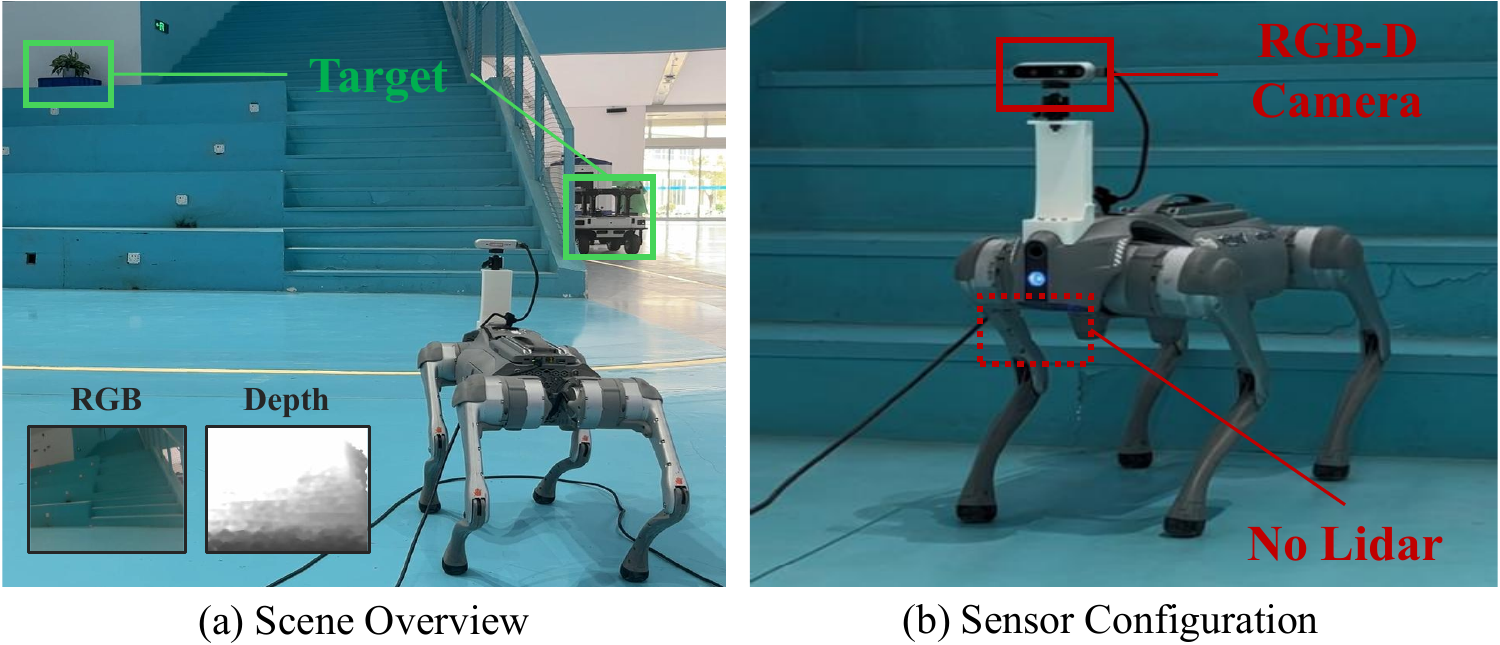}
    \caption{\textbf{Real-world experimental setup.} (a) Multi-floor scenario with target objects located on different floors and RGB-D egocentric observations from the robot. (b) Our quadruped robot equipped with vision-only sensing capabilities for multi-floor navigation tasks.}
\label{fig:kitti}
\vspace{-0.2cm}
\end{figure}

We evaluate performance over eight independent trials per scenario (current-floor, higher-floor, and lower-floor), as shown in Tab.~\ref{tab:reality_stat}. Results highlight the increasing difficulty of cross-floor navigation: stair ascents challenge perception and locomotion, while descents are further complicated by depth sensing and balance maintenance. Despite these challenges, \textbf{\textit{ASCENT}} achieves consistent performance and efficiency, underscoring its robustness in real-world environments.

The experimental setup is shown in Fig.~\ref{fig:kitti}. The robot is equipped with an Intel RealSense D435i for RGB-D perception. High-level planning and reasoning are executed externally: the core algorithm runs on a laptop with an RTX 2060 GPU connected via Ethernet, while LLM queries are offloaded to a remote server with dual RTX 3090 GPUs over a wireless link. Low-level locomotion is controlled by a pre-trained PointNav policy, as in simulation, ensuring stable execution while \textbf{\textit{ASCENT}} focuses on high-level decision-making.

\subsection{Qualitative Analysis} 
\label{sec:qualitative}

Beyond the numerical metrics, we now provide a qualitative analysis to visually demonstrate how our \textbf{\textit{ASCENT}} framework minimizes the dual objective of $c_{\mathrm{expl}}$ and $c_{\mathrm{goal}}$.

\begin{figure}[ht]
    \centering
    \includegraphics[width=\linewidth]{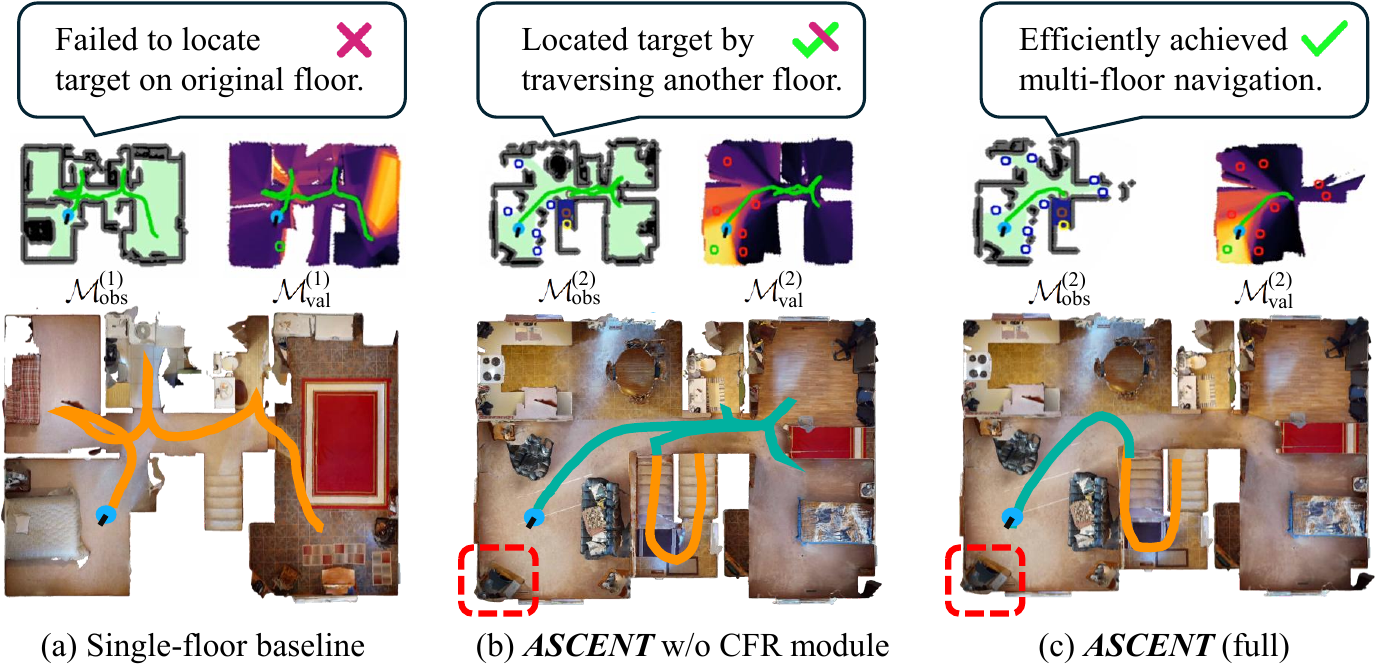}
    \caption{\textbf{Qualitative analysis in multi-floor environments.} We compare the navigation trajectories of different configurations. (a) the single-floor baseline, (b) a variant of our method without the Coarse-to-Fine Reasoning (CFR) module, and (c) the complete \textbf{\textit{ASCENT}}. Overall, \textbf{\textit{ASCENT}} achieves the best balance of success and efficiency.}
\label{fig:vis_traj}
\vspace{-0.2cm}
\end{figure}

Fig. \ref{fig:vis_traj} illustrates the navigation behavior of different methods in a multi-floor scenario. The scenario involves a target object located on the second floor, with the agent starting on the first floor. Panel (a) shows a single-floor baseline that fails to locate the stairway, resulting in mission failure. Panel (b) illustrates a variant of our method without the Coarse-to-Fine Reasoning module. While it successfully navigates to the second floor and reaches the target, it takes an inefficient path. This highlights the importance of our high-level reasoning for reducing  $c_{\mathrm{goal}}$. Panel (c) shows that our \textbf{\textit{ASCENT}} method not only supports multi-floor navigation but also selects a nearly optimal path to the target upon the second floor, achieving high SR while reducing $c_{\mathrm{expl}}$ with superior path efficiency.

\subsection{Robustness Analysis}

\begin{table}[t] 
\centering 
\caption{\revise{\textbf{Recovery analysis for cross-floor episodes.} Recovery Rate is the percentage of episodes with successful failure recovery. Recovery Cost denotes extra timesteps from recovery operations.}}
\resizebox{\linewidth}{!}{
\begin{tabular}{lcccc} 
\toprule 
\textbf{Recovery Type} & \textbf{Failure Type} & \textbf{Percentage (\%)} & \textbf{Recovery Rate (\%)} & \textbf{Avg. Cost (steps)} \\  
\midrule 
No Recovery & Success (no failure) & 47.0 & - & - \\
\midrule 
\multirow{2}{*}{Backtrack} & Stair misidentification & 4.3 & 20.0 & 43 \\ 
& Incorrect floor transition & 12.0 & 14.3 & 198 \\ 
\midrule 
\multirow{2}{*}{Replan} 
& Stuck on stair & 3.4 & 75.0 & 35 \\ 
& Stair detection failure & 33.3 & 23.1 & 12 \\ 
\bottomrule 
\end{tabular}
}
\label{tab:recovery}
\vspace{-0.2cm} 
\end{table}

\noindent \revise{\textbf{Recovery Analysis.} To evaluate multi-floor robustness, we analyze failure patterns across cross-floor episodes on 3 HM3D scenarios, as shown in Table~\ref{tab:recovery}. There are 47.0\% cross-floor episodes succeed without recovery, validating our feasibility checks. For failures, we implement two recovery strategies: Replan recovery handles transient issues (detection failure: 33.3\%, stuck: 3.4\%) via local corrections (12-35 steps), while Backtrack recovery addresses fundamental errors (misidentification: 4.3\%, wrong floor: 12.0\%) through global retrying (43-198 steps). This dual recovery mechanism maintains robustness across diverse failure cross-floor cases.}

\begin{figure}[t]
    \centering
    \includegraphics[width=\linewidth]{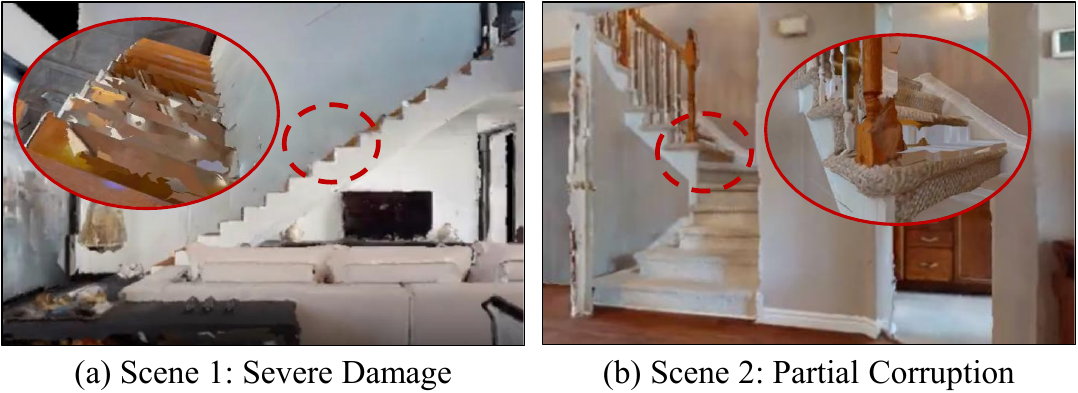}
    \caption{\revise{\textbf{Stress test on visual-geometric mismatch.} We test on two HM3D scenes with corrupted stair geometry where appearance suggests traversability but navmesh damage creates impassable regions. }}
\label{fig:stress_test}
\vspace{-0.2cm}
\end{figure}

\noindent \revise{\textbf{Stress Test.} To evaluate robustness under extreme conditions, we identified two HM3D scenes with naturally occurring navmesh corruption where visual appearance suggests traversability but geometric damage creates impassable regions. As shown in Fig.~\ref{fig:stress_test}, for the cross-floor episodes of these scenes, Scene 1 (severe damage) achieves only 12.5\% SR as extensive corruption blocks most traversable paths. Scene 2 (partial corruption) achieves 36.0\% SR as partial damage allows exploring alternative trajectories. All failures occur at stair traversal, indicating our system is vulnerable when visual perception contradicts geometric reality.}

\section{LIMITATIONS}
\label{sec:limitations}

While OGN assumes static environments, during real-world deployment we have observed that moving obstacles like pedestrians can compromise obstacle mapping accuracy and navigation performance. Future work will integrate adaptive re-planning modules for dynamic scenarios. Additionally, our method is designed for normal staircases and may not perform well on spiral or irregular stairs. Finally, while HM3D and MP3D are widely-used multi-floor benchmarks, future work could explore larger-scale buildings or scenarios.
\section{CONCLUSION}
\label{sec:conclusion}
We presented \textit{\textbf{ASCENT}}, a floor-aware ZS-OGN framework that addresses the limitations of existing methods in online multi-floor navigation. Our approach enables multi-floor planning and context-aware exploration without requiring task-specific training. Experimental results on HM3D and MP3D benchmarks show that our method outperforms state-of-the-art zero-shot methods. We further validated its real-world applicability through deployment on a quadruped robot in unseen environments.
This work contributes to the field by offering a training-free solution for online multi-floor navigation.


\bibliographystyle{IEEEtran} 
\bibliography{IEEEexample}

@article{batra2020objectnav,
    title={Objectnav revisited: On evaluation of embodied agents navigating to objects},
    author={Batra, Dhruv and Gokaslan, Aaron and Kembhavi, Aniruddha and Maksymets, Oleksandr and Mottaghi, Roozbeh and Savva, Manolis and Toshev, Alexander and Wijmans, Erik},
    journal={arXiv preprint arXiv:2006.13171},
    year={2020}
}

@article{ramakrishnan2021habitat,
    title = {Habitat-matterport 3d dataset (hm3d): 1000 large-scale 3d environments for embodied ai},
    author = {Ramakrishnan, Santhosh K and Gokaslan, Aaron and Wijmans, Erik and Maksymets, Oleksandr and Clegg, Alex and Turner, John and Undersander, Eric and Galuba, Wojciech and Westbury, Andrew and Chang, Angel X and others},
    journal = {arXiv preprint arXiv:2109.08238},
    year = {2021}
}

@article{chang2017matterport3d,
    title = {Matterport3d: Learning from rgb-d data in indoor environments},
    author = {Chang, Angel and Dai, Angela and Funkhouser, Thomas and Halber, Maciej and Niessner, Matthias and Savva, Manolis and Song, Shuran and Zeng, Andy and Zhang, Yinda},
    journal = {arXiv preprint arXiv:1709.06158},
    year = {2017}
}

@article{werby23hovsg,
Author = {Abdelrhman Werby and Chenguang Huang and Martin Büchner and Abhinav Valada and Wolfram Burgard},
Title = {Hierarchical Open-Vocabulary 3D Scene Graphs for Language-Grounded Robot Navigation},
Year = {2024},
journal = {Robotics: Science and Systems},
}

@article{chung2024nv,
  title={NV-LIOM: LiDAR-inertial odometry and mapping using normal vectors towards robust SLAM in multifloor environments},
  author={Chung, Dongha and Kim, Jinwhan},
  journal={IEEE Robotics and Automation Letters},
  year={2024},
  publisher={IEEE}
}

@article{kim2024development,
  title={Development of an indoor delivery mobile robot for a multi-floor environment},
  author={Kim, Taejin and Kang, Gyuree and Lee, Daegyu and Shim, D Hyunchul},
  journal={IEEE Access},
  volume={12},
  pages={45202--45215},
  year={2024},
  publisher={IEEE}
}

@article{kuang2024openfmnav,
    title={Openfmnav: Towards open-set zero-shot object navigation via vision-language foundation models},
    author={Kuang, Yuxuan and Lin, Hai and Jiang, Meng},
    journal={arXiv preprint arXiv:2402.10670},
    year={2024}
}

@inproceedings{radford2021learning,
    title={Learning transferable visual models from natural language supervision},
    author={Radford, Alec and Kim, Jong Wook and Hallacy, Chris and Ramesh, Aditya and Goh, Gabriel and Agarwal, Sandhini and Sastry, Girish and Askell, Amanda and Mishkin, Pamela and Clark, Jack and others},
    booktitle={International Conference on Machine Learning},
    pages={8748--8763},
    year={2021},
    organization={PmLR}
}

@inproceedings{li2023blip2,
    title={Blip-2: Bootstrapping language-image pre-training with frozen image encoders and large language models},
    author={Junnan Li and Dongxu Li and Silvio Savarese and Steven Hoi},
    booktitle={International Conference on Machine Learning},
    pages={19730-19742},
    year={2023},
    organization={PMLR}
}

@article{zhang2023llama,
    title={Llama-adapter: Efficient fine-tuning of language models with zero-init attention},
    author={Zhang, Renrui and Han, Jiaming and Liu, Chris and Gao, Peng and Zhou, Aojun and Hu, Xiangfei and Yan, Shilin and Lu, Pan and Li, Hongsheng and Qiao, Yu},
    journal={arXiv preprint arXiv:2303.16199},
    year={2023}
}

@article{liu2023visual,
    title={Visual instruction tuning},
    author={Liu, Haotian and Li, Chunyuan and Wu, Qingyang and Lee, Yong Jae},
    journal={Advances in neural information processing systems},
    volume={36},
    pages={34892--34916},
    year={2023}
}

@article{liu2019roberta,
    title={Roberta: A robustly optimized bert pretraining approach},
    author={Liu, Yinhan and Ott, Myle and Goyal, Naman and Du, Jingfei and Joshi, Mandar and Chen, Danqi and Levy, Omer and Lewis, Mike and Zettlemoyer, Luke and Stoyanov, Veselin},
    journal={arXiv preprint arXiv:1907.11692},
    year={2019}
}

@article{bai2023qwen,
    title={Qwen-VL: A Versatile Vision-Language Model for Understanding, Localization},
    author={Bai, Jinze and Bai, Shuai and Yang, Shusheng and Wang, Shijie and Tan, Sinan and Wang, Peng and Lin, Junyang and Zhou, Chang and Zhou, Jingren},
    journal={arXiv preprint arXiv:2308.12966},
    year={2023}
}

@article{team2024qwen2,
  title={Qwen2 technical report},
  author={Team, Qwen},
  journal={arXiv preprint arXiv:2407.10671},
  year={2024}
}

@article{yang2024qwen2,
    title={Qwen2.5 technical report},
    author={Yang, An and Yang, Baosong and Zhang, Beichen and Hui, Binyuan and Zheng, Bo and Yu, Bowen and Li, Chengyuan and Liu, Dayiheng and Huang, Fei and Wei, Haoran and others},
    journal={arXiv preprint arXiv:2412.15115},
    year={2024}
}

@article{achiam2023gpt,
    title={Gpt-4 technical report},
    author={Achiam, Josh and Adler, Steven and Agarwal, Sandhini and Ahmad, Lama and Akkaya, Ilge and Aleman, Florencia Leoni and Almeida, Diogo and Altenschmidt, Janko and Altman, Sam and Anadkat, Shyamal and others},
    journal={arXiv preprint arXiv:2303.08774},
    year={2023}
}

@article{zhang2024multi,
    title={Multi-Floor Zero-Shot Object Navigation Policy},
    author={Zhang, Lingfeng and Wang, Hao and Xiao, Erjia and Zhang, Xinyao and Zhang, Qiang and Jiang, Zixuan and Xu, Renjing},
    journal={arXiv preprint arXiv:2409.10906},
    year={2024}
}

@article{majumdar2022zson,
    title={Zson: Zero-shot object-goal navigation using multimodal goal embeddings},
    author={Majumdar, Arjun and Aggarwal, Gunjan and Devnani, Bhavika and Hoffman, Judy and Batra, Dhruv},
    journal={Advances in Neural Information Processing Systems},
    volume={35},
    pages={32340--32352},
    year={2022}
}

@inproceedings{yu2023l3mvn,
    title={L3mvn: Leveraging large language models for visual target navigation},
    author={Yu, Bangguo and Kasaei, Hamidreza and Cao, Ming},
    booktitle={IEEE/RSJ International Conference on Intelligent Robots and Systems},
    pages={3554--3560},
    year={2023},
}

@inproceedings{yokoyama2024vlfm,
    title={Vlfm: Vision-language frontier maps for zero-shot semantic navigation},
    author={Yokoyama, Naoki and Ha, Sehoon and Batra, Dhruv and Wang, Jiuguang and Bucher, Bernadette},
    booktitle={IEEE International Conference on Robotics and Automation},
    pages={42--48},
    year={2024},
}

@inproceedings{longinstructnav,
    title={InstructNav: Zero-shot System for Generic Instruction Navigation in Unexplored Environment},
    author={Long, Yuxing and Cai, Wenzhe and Wang, Hongcheng and Zhan, Guanqi and Dong, Hao},
    booktitle={Conference on Robot Learning},
    year={2024}
}

@inproceedings{cai2024bridging,
    title={Bridging zero-shot object navigation and foundation models through pixel-guided navigation skill},
    author={Cai, Wenzhe and Huang, Siyuan and Cheng, Guangran and Long, Yuxing and Gao, Peng and Sun, Changyin and Dong, Hao},
    booktitle={IEEE International Conference on Robotics and Automation},
    pages={5228--5234},
    year={2024},
}

@inproceedings{ramrakhya2023pirlnav,
    title={Pirlnav: Pretraining with imitation and rl finetuning for objectnav},
    author={Ramrakhya, Ram and Batra, Dhruv and Wijmans, Erik and Das, Abhishek},
    booktitle={IEEE/CVF Conference on Computer Vision and Pattern Recognition},
    pages={17896--17906},
    year={2023}
}

@article{yin2025sg,
    title={SG-Nav: Online 3D Scene Graph Prompting for LLM-based Zero-shot Object Navigation},
    author={Yin, Hang and Xu, Xiuwei and Wu, Zhenyu and Zhou, Jie and Lu, Jiwen},
    journal={Advances in Neural Information Processing Systems},
    volume={37},
    pages={5285--5307},
    year={2024}
}

@inproceedings{wasserman2024exploitation,
    title={Exploitation-guided exploration for semantic embodied navigation},
    author={Wasserman, Justin and Chowdhary, Girish and Gupta, Abhinav and Jain, Unnat},
    booktitle={IEEE International Conference on Robotics and Automation},
    pages={2901--2908},
    year={2024},
}

@inproceedings{savva2019habitat,
    title={Habitat: A platform for embodied ai research},
    author={Savva, Manolis and Kadian, Abhishek and Maksymets, Oleksandr and Zhao, Yili and Wijmans, Erik and Jain, Bhavana and Straub, Julian and Liu, Jia and Koltun, Vladlen and Malik, Jitendra and others},
    booktitle={IEEE/CVF International Conference on Computer Vision},
    pages={9339--9347},
    year={2019}
}

@inproceedings{chen2023object,
  title={Object goal navigation with recursive implicit maps},
  author={Chen, Shizhe and Chabal, Thomas and Laptev, Ivan and Schmid, Cordelia},
  booktitle={2023 IEEE/RSJ International Conference on Intelligent Robots and Systems (IROS)},
  pages={7089--7096},
  year={2023},
  organization={IEEE}
}

@article{jiang2018rednet,
  title={Rednet: Residual encoder-decoder network for indoor rgb-d semantic segmentation},
  author={Jiang, Jindong and Zheng, Lunan and Luo, Fei and Zhang, Zhijun},
  journal={arXiv preprint arXiv:1806.01054},
  year={2018}
}

@inproceedings{liu2024grounding,
  title={Grounding dino: Marrying dino with grounded pre-training for open-set object detection},
  author={Liu, Shilong and Zeng, Zhaoyang and Ren, Tianhe and Li, Feng and Zhang, Hao and Yang, Jie and Jiang, Qing and Li, Chunyuan and Yang, Jianwei and Su, Hang and others},
  booktitle={European Conference on Computer Vision},
  pages={38--55},
  year={2024},
  organization={Springer}
}

@article{peng2024d,
  title={D-FINE: redefine regression Task in DETRs as Fine-grained distribution refinement},
  author={Peng, Yansong and Li, Hebei and Wu, Peixi and Zhang, Yueyi and Sun, Xiaoyan and Wu, Feng},
  journal={arXiv preprint arXiv:2410.13842},
  year={2024}
}

@article{zhang2023faster,
  title={Faster segment anything: Towards lightweight sam for mobile applications},
  author={Zhang, Chaoning and Han, Dongshen and Qiao, Yu and Kim, Jung Uk and Bae, Sung-Ho and Lee, Seungkyu and Hong, Choong Seon},
  journal={arXiv preprint arXiv:2306.14289},
  year={2023}
}

@article{zhou2017places,
  title={Places: A 10 million image database for scene recognition},
  author={Zhou, Bolei and Lapedriza, Agata and Khosla, Aditya and Oliva, Aude and Torralba, Antonio},
  journal={IEEE transactions on pattern analysis and machine intelligence},
  volume={40},
  number={6},
  pages={1452--1464},
  year={2017},
  publisher={IEEE}
}

@inproceedings{zhang2024recognize,
  title={Recognize anything: A strong image tagging model},
  author={Zhang, Youcai and Huang, Xinyu and Ma, Jinyu and Li, Zhaoyang and Luo, Zhaochuan and Xie, Yanchun and Qin, Yuzhuo and Luo, Tong and Li, Yaqian and Liu, Shilong and others},
  booktitle={Proceedings of the IEEE/CVF Conference on Computer Vision and Pattern Recognition},
  pages={1724--1732},
  year={2024}
}

@misc{habitatchallenge2022,
  title        = "{Habitat Challenge 2022}",
  author       = "Karmesh Yadav and Santhosh Kumar Ramakrishnan and John Turner and 
                  Aaron Gokaslan and Oleksandr Maksymets and Rishabh Jain and 
                  Ram Ramrakhya and Angel X. Chang and Alexander Clegg and 
                  Manolis Savva and Eric Undersander and Devendra Singh Chaplot and 
                  Dhruv Batra",
  howpublished = "\url{https://aihabitat.org/challenge/2022/ }",
  year         = "2022"
}

@article{hamilton2017inductive,
  title={Inductive representation learning on large graphs},
  author={Hamilton, Will and Ying, Zhitao and Leskovec, Jure},
  journal={Advances in neural information processing systems},
  volume={30},
  year={2017}
}

@article{yoo2024commonsense,
  title={Commonsense-aware object value graph for object goal navigation},
  author={Yoo, Hwiyeon and Choi, Yunho and Park, Jeongho and Oh, Songhwai},
  journal={IEEE Robotics and Automation Letters},
  volume={9},
  number={5},
  pages={4423--4430},
  year={2024},
  publisher={IEEE}
}

@article{guo2024object,
  title={An object-driven navigation strategy based on active perception and semantic association},
  author={Guo, Yu and Sun, Jinsheng and Zhang, Ruiheng and Jiang, Zhiqi and Mi, Zhenqiang and Yao, Chao and Ban, Xiaojuan and Obaidat, Mohammad S},
  journal={IEEE Robotics and Automation Letters},
  volume={9},
  number={8},
  pages={7110--7117},
  year={2024},
  publisher={IEEE}
}

@article{zhu2025strive,
  title={Strive: Structured representation integrating vlm reasoning for efficient object navigation},
  author={Zhu, Haokun and Li, Zongtai and Liu, Zhixuan and Wang, Wenshan and Zhang, Ji and Francis, Jonathan and Oh, Jean},
  journal={arXiv preprint arXiv:2505.06729},
  year={2025}
}

@article{zhang2025apexnav,
  title={ApexNav: An Adaptive Exploration Strategy for Zero-Shot Object Navigation with Target-centric Semantic Fusion},
  author={Zhang, Mingjie and Du, Yuheng and Wu, Chengkai and Zhou, Jinni and Qi, Zhenchao and Ma, Jun and Zhou, Boyu},
  journal={arXiv preprint arXiv:2504.14478},
  year={2025}
}

@article{liu2024deepseek,
  title={Deepseek-v3 technical report},
  author={Liu, Aixin and Feng, Bei and Xue, Bing and Wang, Bingxuan and Wu, Bochao and Lu, Chengda and Zhao, Chenggang and Deng, Chengqi and Zhang, Chenyu and Ruan, Chong and others},
  journal={arXiv preprint arXiv:2412.19437},
  year={2024}
}


\definecolor{brightpink}{rgb}{1.0, 0.0, 0.5}
\newcommand{\myPink}[1]{\textcolor{brightpink}{#1}}

\section*{Appendix}
\startcontents[appendices]
\printcontents[appendices]{l}{1}{\setcounter{tocdepth}{3}}

\setcounter{table}{0}
\setcounter{figure}{0}
\setcounter{section}{0}

\renewcommand{\thetable}{A\arabic{table}}
\renewcommand{\thefigure}{A\arabic{figure}}
\renewcommand{\thesection}{A\arabic{section}}

\section{Additional Implementation Details}
In this section, we elaborate on additional experimental settings and implementation details to facilitate the reproduction of the approach proposed in this work.

\subsection{Problem Formulation}
We tackle the OGN task, where a robot $r$ starts from an initial position $p_0 \in \mathbb{R}^{3}$ in an unexplored environment. The robot only has access to an egocentric RGB-D camera and an odometry sensor that provides position $p$ relative to $p_0$. At each timestep $t$, the robot selects discrete actions $a_t \in \mathcal{A}$ to generate a path $\tau$ towards discovering any instance of a target object category $\mathcal{O}$. The action space $\mathcal{A}$ consists of: MOVE FORWARD (0.25m), TURN LEFT ($30^{\circ}$), TURN RIGHT ($30^{\circ}$), LOOK UP ($30^{\circ}$), LOOK DOWN ($30^{\circ}$), and STOP. The timestep limit $T$ is $500$ steps. Following the oracle-visibility criterion of OGN evaluation~\cite{batra2020objectnav}, an episode is successful if:
\begin{equation}
    \tau(T) \in \big\{ p \in \mathbb{R}^{3} \ | d_v(p, \mathcal{O}) \leq 1\,\text{m} \land a_T = \text{STOP} \big\}
\end{equation}

\noindent where $d_v(p, \mathcal{O}) \leq 1\,\text{m}$ indicates target $\mathcal{O}$ is within one-meter distance and visible via camera rotation from position $p$.

\subsection{Motivation of Prior Knowledge}
\label{sec:pk}
Before navigation, we extract statistical priors from the training sets of corresponding large-scale datasets, including \textit{floor-level} and \textit{area-level} object distributions. These priors serve two key purposes: First, they reflect dataset-specific patterns that learning-based methods may implicitly encode, which provides a structured reference for our zero-shot framework to align with expected object locations. Second, in the absence of global map knowledge, they offer valuable guidance for inter-floor reasoning, helping the agent prioritize unexplored floors or areas where target objects are more likely to appear based on environmental context.

Incorporating prior knowledge into our system is designed to enhance performance, although it is not essential for the core functionality. The Floor Prior Knowledge is derived from the training data of the HM3D and MP3D datasets, which are tailored to match the characteristics of their respective test datasets. This approach improves the spatial understanding necessary for multi-floor navigation tasks. The Scene Prior Knowledge is gleaned from the statistical properties of the HM3D dataset, which enriches the LLM's comprehension of common scene elements and their typical configurations.

As discussed in Tab.~\ref{tab:ablation}, we have already explored the influence of this prior knowledge. The LLM possesses inherent capabilities that enable it to adapt and deliver sensible outputs even in the absence of prior knowledge. Nevertheless, when augmented with the statistical insights gleaned from the datasets, the LLM operates more efficiently and with enhanced accuracy, thereby offering a marginal improvement in performance.










\subsection{Vertical Localization Constraints}
\label{sec:appendix_multifloor_design}

In real-world settings, precise vertical localization remains challenging due to sensor limitations and environmental factors. Widely used benchmarks such as the Habitat Navigation Challenge~\cite{habitatchallenge2022} constrain agents to GPS+Compass sensors that provide only 2D horizontal localization, ensuring fair comparison across methods and reflecting practical sim-to-real constraints.

IMU-derived height estimates are prone to drift and lack precision during dynamic movements like stair climbing. Visual SLAM also struggles in texture-poor environments such as staircases, where feature extraction becomes unreliable. Even if vertical coordinates were available, determining floor levels would still require knowledge of building-specific floor heights. However, prior information is typically unknown in unseen environments. Small sensor errors could lead to incorrect floor classification and planning failures.

Instead of relying on noisy or unavailable vertical data, we propose a vision-based, floor-aware modeling strategy that infers inter-floor transitions directly from scene semantics and depth anomalies. This approach offers a more robust and realistic solution for multi-floor navigation under RGB-D and 2D odometry constraints.

\subsection{Details of Coarse-Grained Frontier Proposal} 
\label{sec:appendix_coarse}
Frontiers are defined as the boundary midpoints between explored and unexplored regions. For each detected frontier, the corresponding RGB image at that timestep is processed to generate a structured scene description using a scene classification model and an image tagging model. These descriptions capture both room types (\eg, “living room”) and associated objects (\eg, “sofa”, “TV”), forming a rich contextual basis for LLM reasoning.

To further improve efficiency and reduce redundancy, we apply a visual similarity filter to remove the frontiers whose representative images stay below a threshold $\theta$ on structural overlap. This design prioritizes frontiers based on scene semantics and spatial diversity:
\begin{equation}
\mathcal{F}_{\mathrm{coarse}} = \left\{ f \in \mathcal{F}_{\mathrm{boundary}} \,\middle|\, \text{SSIM}(f, \mathcal{F}_{\mathrm{cache}}) < \tau_\mathrm{ssim} \right\},
\end{equation}
where $ \mathcal{F}_{\mathrm{boundary}} $ denotes frontier boundary points between explored and unexplored regions, and $ \mathcal{F}_{\mathrm{cache}} $ stores previously evaluated frontiers. The Structural Similarity Index Measure (SSIM) quantifies perceptual similarity between image pairs by comparing luminance, contrast, and structure:
\begin{equation}
\text{SSIM}(x, y) = \frac{(2\mu_x\mu_y + c_1)(2\sigma_{xy} + c_2)}{(\mu_x^2 + \mu_y^2 + c_1)(\sigma_x^2 + \sigma_y^2 + c_2)},
\end{equation}
where $ \mu $, $ \sigma $, and $ \sigma_{xy} $ denote local means, standard deviations, and cross-covariance of image intensities, respectively; $ c_1 $ and $ c_2 $ are small constants to stabilize division. We use this metric to filter out visually redundant frontiers, ensuring spatial diversity in the candidate set.

\section{Details of Multi-Floor Navigation}

The Multi-Floor Spatial Abstraction module is a key innovation of our work, enabling floor awareness for the agent and serving as a plug-and-play module for navigation tasks. This section details its working principle and navigation impact through both textual description and visual analysis.

\subsection{Motivation of Multi-Floor Navigation}
\label{sec:appendix_multifloor_statics}

\begin{figure}[ht]
    \centering
    \includegraphics[width=\linewidth]{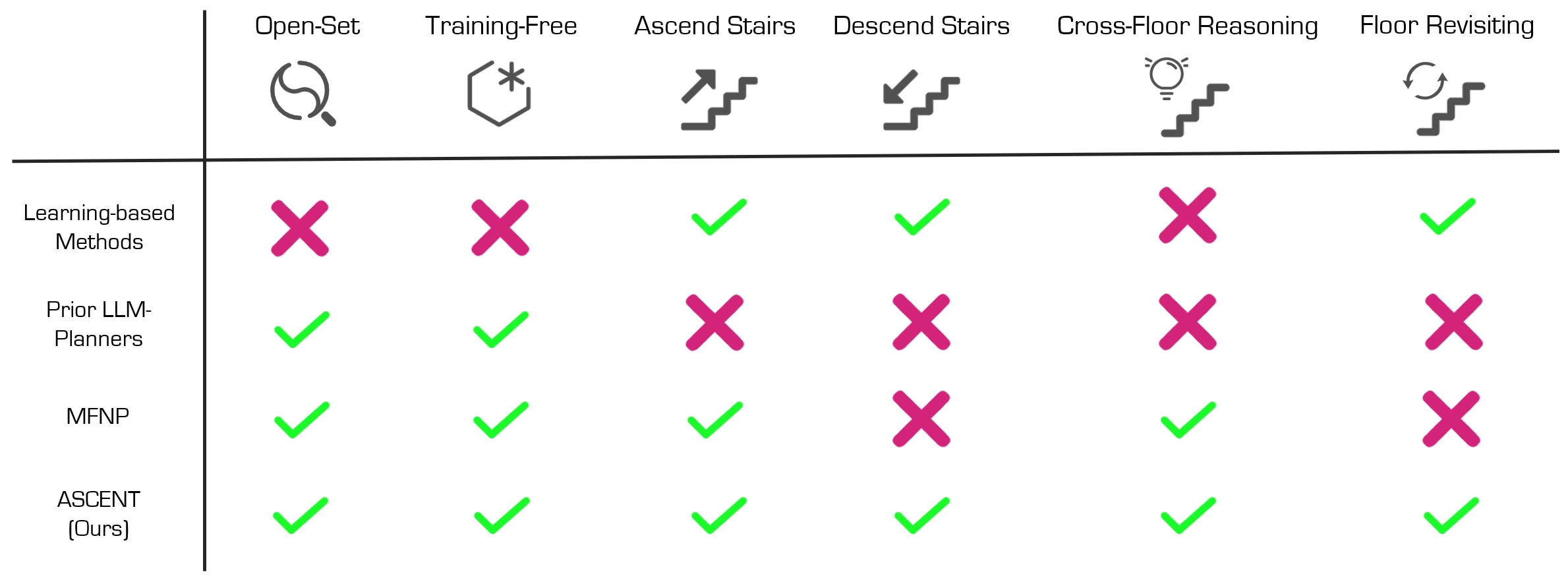}
    \caption{Capability comparisons of navigation methods.}
    \label{fig:comp_abli}
    \vspace{-0.2cm}
\end{figure}

Currently, most navigation systems overlook the cross-floor capabilities required for such tasks, leading to failures in these episodes and creating bottlenecks in success rates. As shown in Fig. \ref{fig:comp_abli}, previous LLM-Planners, with the exception of MFNP, often lack multi-floor navigation capabilities. They either fail to move between floors or do not recognize when a floor change has occurred, resulting in semantic map confusion or overlap. While MFNP can detect upstairs transitions through image recognition and perform cross-floor detection, it struggles with downstairs transitions (since downstairs stairs are not directly visible from the default horizontal perspective). This limits its cross-floor capabilities. Additionally, MFNP treats stair entrances as obstacles to prevent floor transition failures, which inadvertently prevents re-visiting floors. Although learning-based methods can handle multi-floor navigation, they lack interpretability in cross-floor reasoning and planning, and they fail to generalize to open-vocabulary detection tasks. This has inspired us to explore adaptive multi-floor navigation algorithms that address these limitations.

\subsection{Cases Analysis for Stair Ascending}
When the target is located on a higher floor, the agent must navigate upstairs. As shown in Fig.~\ref{fig:supp_upstair}, the process for upstairs navigation is as follows:

\begin{figure*}[h!]
    \centering
    \includegraphics[width=\linewidth]{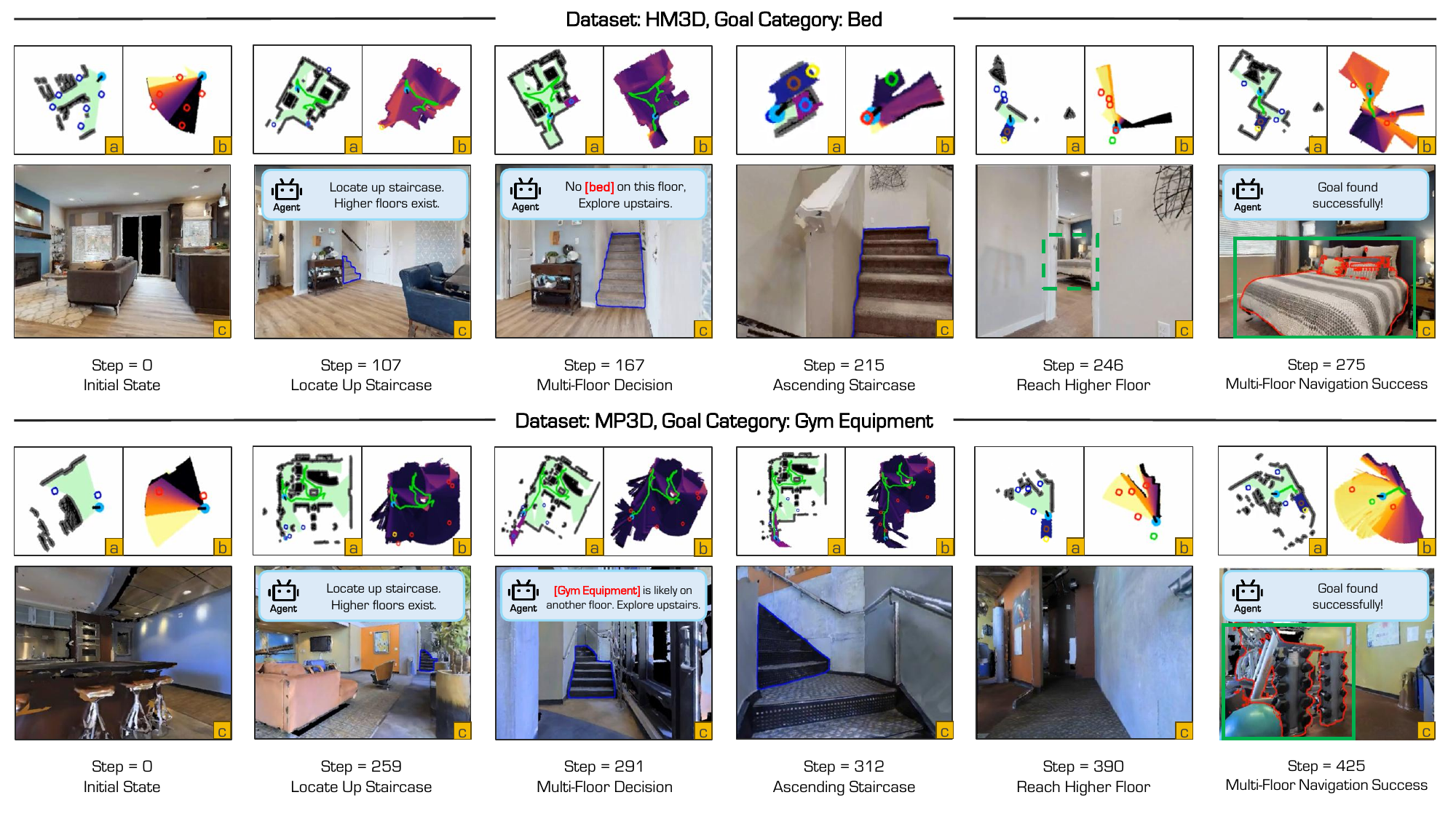}
    \vspace{-0.5cm}
    \caption{\textbf{Cases Analysis for Stair Ascending.} \textbf{Top row:} After traversing the current floor, the agent makes a multi-floor decision to navigate upstairs and successfully find the goal on a higher floor in the HM3D dataset. \textbf{Bottom row:} The agent actively infers that the target is on a higher floor and decides to ascend, successfully navigating to the goal on a higher floor in the MP3D dataset.}
    \label{fig:supp_upstair}
\end{figure*}
\begin{figure*}[h!]
    \centering
    \includegraphics[width=\linewidth]{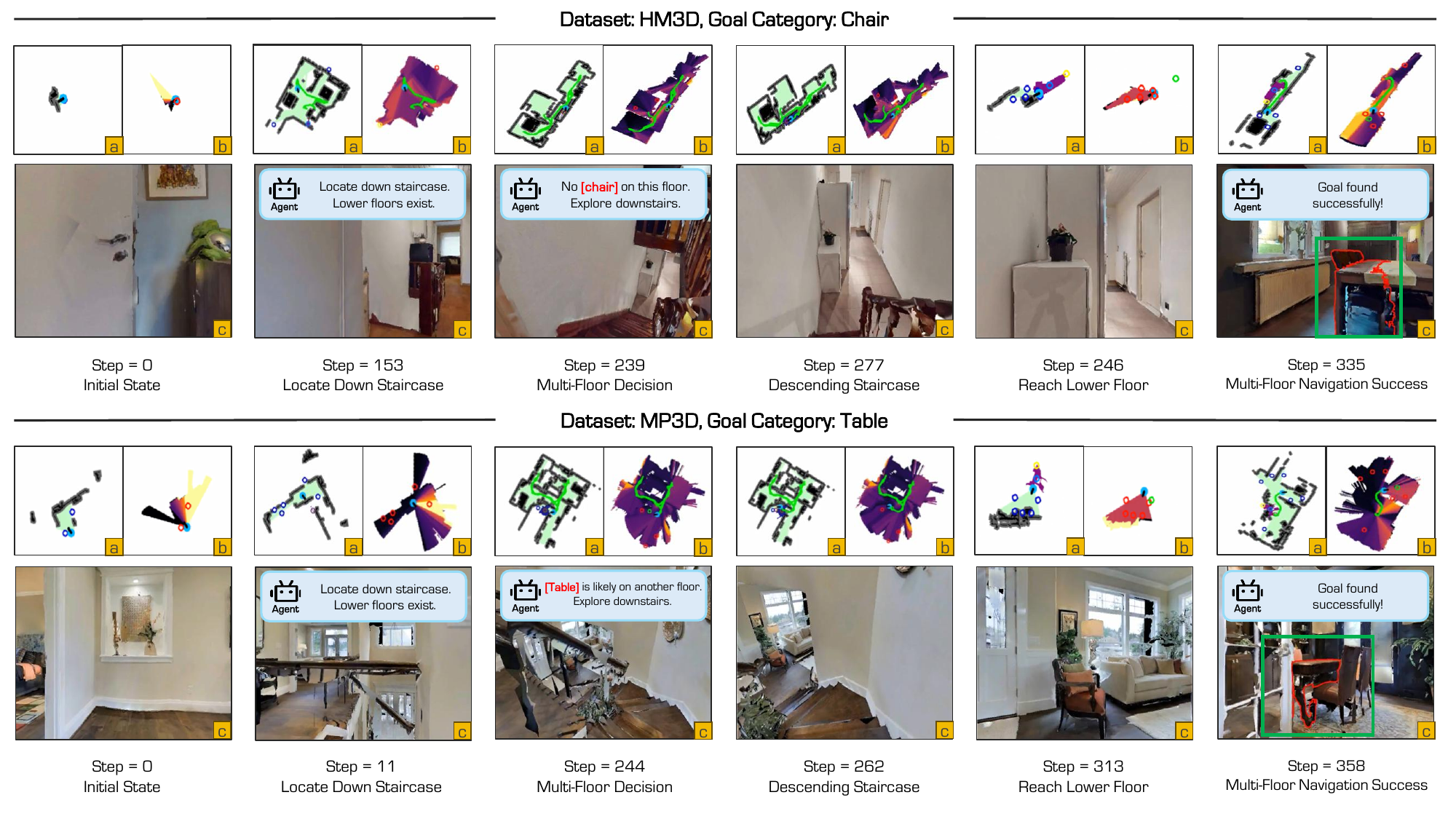}
    \vspace{-0.5cm}
    \caption{\textbf{Cases Analysis for Stair Descending.} \textbf{Top row:} After traversing the current floor, the agent makes a multi-floor decision to navigate downstairs and successfully find the goal on a lower floor in the HM3D dataset. \textbf{Bottom row:} The agent actively infers that the target is on a lower floor and decides to descend, successfully navigating to the goal on a lower floor in the MP3D dataset.}
    \label{fig:supp_downstair}
\end{figure*}

Initially, the agent operates under the assumption of a single floor, using BEV map lists $\mathcal{L}_{obs} = [\mathcal{M}_{obs}^{(1)}]$ and $\mathcal{L}_{val} = [\mathcal{M}_{val}^{(1)}]$, performing intra-floor navigation until detecting an upstairs region through both staircase detection and semantic segmentation. Upon detection, it initializes maps for the higher floor ($\mathcal{M}_{obs}^{(2)}$, $\mathcal{M}_{val}^{(2)}$). 

The agent initiates upstairs transition when either: (1) the current floor exploration completes while higher unexplored floors remain, as shown in the example of the top row in Fig.~\ref{fig:supp_upstair}; or (2) the step count exceeds $\tau_{\mathrm{floor}}$ steps and the time since the last floor decision exceeds $\tau_{\mathrm{period}}$ steps (triggering the LLM's multi-floor decision), as shown in the example of the bottom row in Fig.~\ref{fig:supp_upstair}.

During the transition, the agent treats the centroid of the stair as its initial waypoint and marks the entry point when its centroid enters the stair region. When stair pixels in the upper half image exceed a threshold $\epsilon_{up}$, the agent executes a \lookup action to prevent the pretrained PointNav model from mistaking the stair for an obstacle. After reaching the centroid of the stair, it navigates using dynamically updated waypoints ($\sigma$ meters ahead) as described in Sec.~\ref{sec:method}. The exit point is marked when the centroid fully leaves the stair region. This process ensures smooth and accurate navigation through the stair region.

After ascending the stairs, the agent initializes the navigation stage on the new floor while establishing floor connectivity through bidirectional stair mapping using the BEV maps $\mathcal{M}_{obs}^{(2)}$ and $\mathcal{M}_{val}^{(2)}$. The higher floor's down-stair parameters inherit the lower floor's up-stair geometry with start/end points inverted while preserving the centroid. Navigation then continues with intra-floor rules, periodically invoking the LLM decision when unexplored floors remain in the exploration state.

\subsection{Cases Analysis for Stair Descending}
When the target is located on a lower floor, the agent must navigate downstairs. As shown in Fig.~\ref{fig:supp_downstair}, the process for downstairs navigation is similar to that for ascending stairs, with the following adjustments:

Similarly, the agent operates under the assumption of a single floor, using BEV map lists $\mathcal{L}_{obs} = [\mathcal{M}_{obs}^{(1)}]$ and $\mathcal{L}_{val} = [\mathcal{M}_{val}^{(1)}]$. However, while it detects a down-stair region through depth inversion analysis, it transfer its current maps as ($\mathcal{M}_{obs}^{(2)}$, $\mathcal{M}_{val}^{(2)}$) and initializes maps for the lower floor ($\mathcal{M}_{obs}^{(1)}$, $\mathcal{M}_{val}^{(1)}$). 

The agent initiates downstairs transition when either: (1) the current floor exploration completes while lower unexplored floors remain, as shown in the example of the top row in Fig.~\ref{fig:supp_downstair}; or (2) the step count exceeds $\tau_{\mathrm{floor}}$ steps and the time since the last floor decision exceeds $\tau_{\mathrm{period}}$ steps (triggering the LLM's multi-floor decision), as shown in the example of the bottom row in Fig.~\ref{fig:supp_downstair}.

During the transition, the agent treats the centroid of the stair as its initial waypoint and marks the entry point when its centroid enters the stair region. When stair pixels in the lower half image are fewer than a threshold $\epsilon_{down}$, the agent executes a \lookdown action to prevent losing the view of staircases. When stair pixels in the upper half image exceed a threshold $\epsilon_{up}$, the agent executes a \lookup action to prevent the pretrained PointNav model from mistaking the stair for an obstacle. After reaching the centroid of the stair, it navigates using dynamically updated waypoints ($\sigma$ meters ahead) as described in Sec.~\ref{sec:method}. The exit point is marked when the centroid fully leaves the stair region. This process ensures smooth and accurate navigation through the stair region.

After descending the stairs, the agent initializes the navigation stage on the new floor while establishing floor connectivity through bidirectional stair mapping using the BEV maps $\mathcal{M}_{obs}^{(1)}$ and $\mathcal{M}_{val}^{(1)}$. The lower floor's upstairs parameters inherit the higher floor's down-stair geometry with start/end points inverted while preserving the centroid. Navigation then continues with intra-floor rules, periodically invoking the LLM decision when unexplored floors remain in the exploration state.

\begin{figure*}[t]
    \centering
    \includegraphics[width=\linewidth]{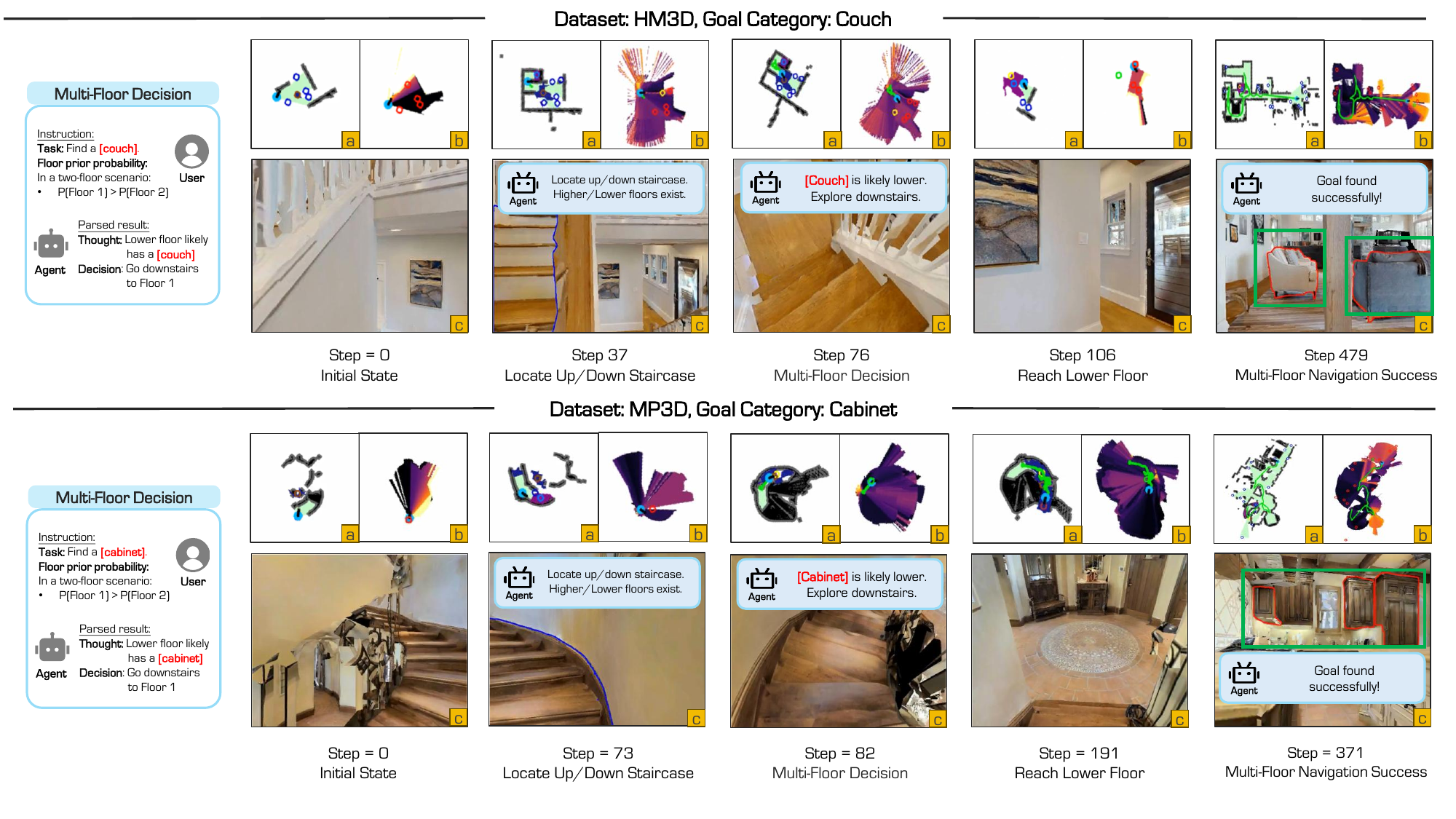}
    \vspace{-0.6cm}
    \caption{\textbf{Cases Analysis for Stairwells:} These examples illustrate the agent's capability to make multi-floor decisions despite limited environmental context. \textbf{Top row:} Starting on a landing between stair flights in the HM3D dataset, the agent leverages prior floor knowledge to make a multi-floor decision, navigate downstairs, and successfully locate the goal on a lower floor. \textbf{Bottom row:} Beginning within the stair area in MP3D, the agent proactively infers that the goal is likely on a lower floor, leading to a multi-floor decision to navigate downwards and successfully reach the goal.}
    \label{fig:supp_stairwell}
    \vspace{-0.2cm}
\end{figure*}
\begin{figure*}[t]
    \centering
    \includegraphics[width=\linewidth]{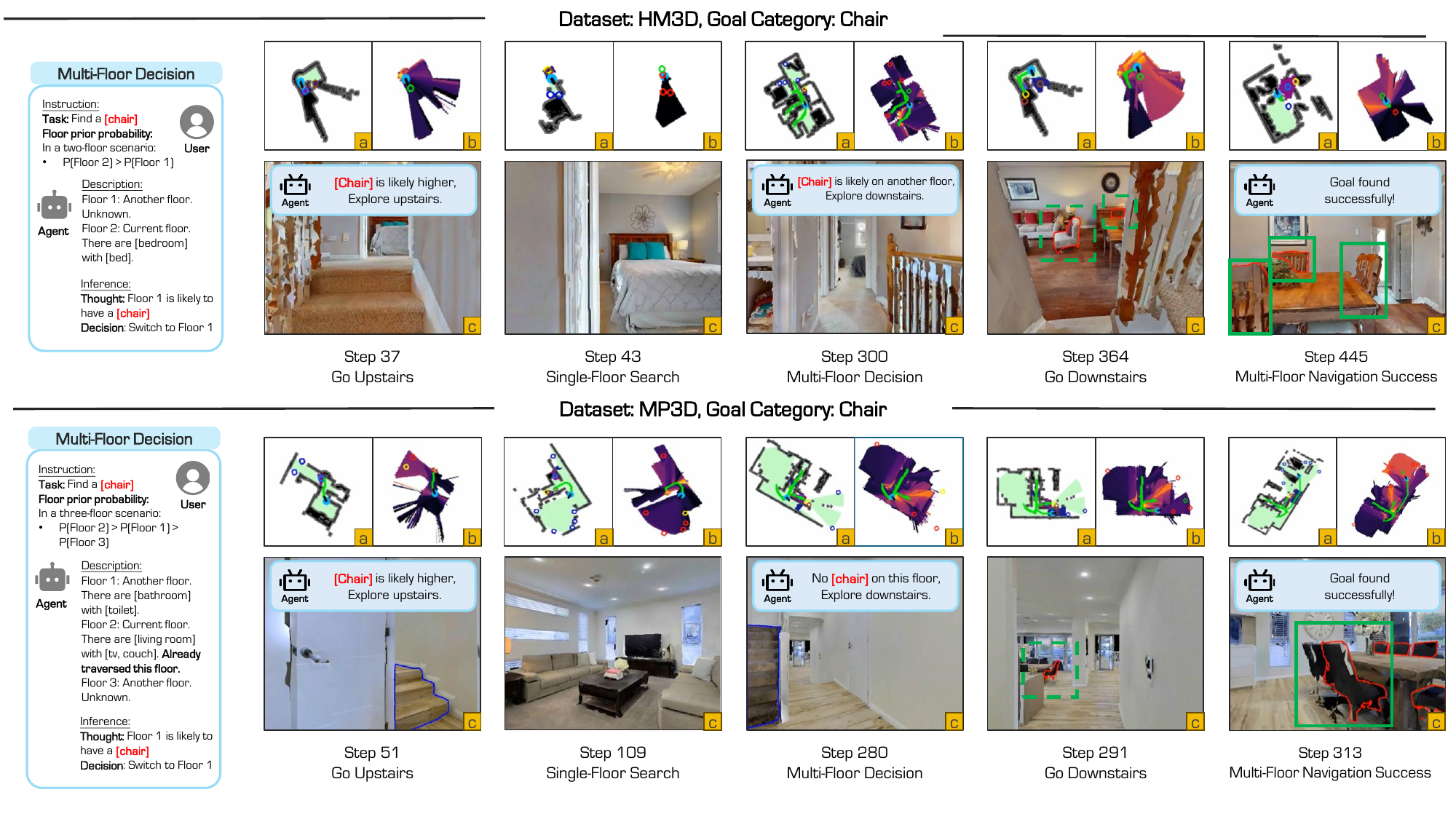}
    \vspace{-0.6cm}
    \caption{\textbf{Cases Analysis for Floor Revisiting:} These examples show the agent revisiting floors in complex multi-floor scenarios, illustrating the robustness of our algorithm. \textbf{Top row:} This example from the HM3D dataset, where the agent initially makes the wrong multi-floor decision but eventually successfully finds the target object through floor revisiting. \textbf{Bottom row:} In an example from the MP3D dataset featuring a scenario with multiple staircases connecting the first and second floors, the agent is still capable of revisiting the first floor and successfully reaching the target.}
    \label{fig:supp_revisit}
    \vspace{-0.2cm}
\end{figure*}

\subsection{Cases Analysis for Stairwells}
The agent may initialize its navigation within a stairwell, either positioned on a landing between stair flights (Fig.~\ref{fig:supp_stairwell}, top) or directly inside the stair section (Fig.~\ref{fig:supp_stairwell}, bottom). In such stairwell initialization scenarios, the agent faces unique challenges compared to conventional floor-based starts. The constrained field of view within stairwells limits the available semantic and geometric information for initial scene understanding. Due to limited initial contextual information of the scenario environment, we incorporate prior floor knowledge with the specific goal into the LLM. When the agent detects both ascending and descending staircases, this prior knowledge informs the LLM's multi-floor decision-making process, enabling probabilistic estimation of the target's vertical location (upstairs vs. downstairs). This optimization significantly improves navigation efficiency. This combined geometric and semantic analysis enables the agent to make informed vertical navigation decisions even from spatially constrained starting positions, demonstrating the robustness of \textbf{\textit{ASCENT}} framework in handling diverse initialization conditions.


\subsection{Cases Analysis for Floor Revisiting}
As shown in Fig.~\ref{fig:supp_revisit}, these cases demonstrate the effectiveness and robustness of our algorithm in multi-floor navigation. 

The top row of examples is from the scene of the HM3D dataset. The agent initially leverages prior floor knowledge to make multi-floor decisions. After an unsuccessful search on the higher floor, the system actively reevaluates its strategy and successfully relocates the target might on the lower floor through a timely floor revisit. This adaptive behavior demonstrates the algorithm's robustness in recovering from initial decision inaccuracies.

The bottom row presents a more complex case with dual staircases connecting the first and second floors from the MP3D dataset. The scene has two staircases between the 1st floor and the $2$nd floor. Following partial exploration of the first floor, the agent prioritizes searching the second floor based on prior probability estimates. After searching for a while on the $2$nd floor but not finding the target object, the agent needs to choose to go up to the $3$rd floor or down to the 1st floor. Because the prior knowledge suggests that the probability of finding the target on the 1st floor is higher than that on the 3rd floor, the agent revisits the $1$st floor from another staircase and successfully reaches the target, which validates both the effectiveness of our probability-driven approach and the algorithm's capability to handle architectural complexities.
\clearpage\clearpage
\section{Detailed Prompt Design}
\label{sec_supp:prompt}

\begin{table}[ht]
    \centering
    \caption{An example of the instruction for single-floor object navigation.}
    \vspace{-0.2cm}
    \renewcommand{\arraystretch}{1.2}
    \setlength{\tabcolsep}{8pt}
    \tcbset{
        colback=gray!10,
        colframe=gray!50,
        arc=2mm,
        boxrule=0.3mm,
        width=\linewidth,  
        left=3pt,          
        right=3pt,         
        top=3pt,           
        bottom=3pt,        
        boxsep=0pt,        
    }
    \begin{tcolorbox}
    \small
    \textbf{\texttt{Task: Select the optimal area based on prior}}\\
    \textbf{\texttt{probabilistic data and environmental context.}}
    
    \vspace{0.2cm}
    
    \texttt{\textbf{Expected Answer Format: JSON}}
    
    \vspace{0.2cm}
    
    \texttt{\textbf{\textcolor{orange}{Example Input:}}}
    
\begin{lstlisting}[
    basicstyle=\ttfamily\scriptsize,
    breaklines=true,
    breakatwhitespace=true,
    columns=flexible,
    keepspaces=true,
    xleftmargin=0pt,
    framexleftmargin=0pt,
    aboveskip=3pt,
    belowskip=3pt
]
{
  "Goal": "toilet",
  "Prior Probabilities between Room Type and 
   Goal Object": [
    "Bathroom": 90.0%, "Bedroom": 10.0%
  ],
  "Area Descriptions": [
    "Area 1": "a bathroom containing objects: 
               shower, towel",
    "Area 2": "a bedroom containing objects: 
               bed, nightstand",
    "Area 3": "a garage containing objects: car"
  ]
}
\end{lstlisting}

    \texttt{\textbf{\textcolor{orange}{Example Response:}}}
    
\begin{lstlisting}[
    basicstyle=\ttfamily\scriptsize,
    breaklines=true,
    breakatwhitespace=true,
    columns=flexible,
    keepspaces=true,
    xleftmargin=0pt,
    framexleftmargin=0pt,
    aboveskip=3pt,
    belowskip=3pt
]
{
  "Index": "1",
  "Reason": "Shower and towel in Bathroom indicate 
             toilet location, with high probability 
             (90.0%)."
}
\end{lstlisting}

    \texttt{\textbf{\textcolor{brown}{Prompted Input:}}}
    
\begin{lstlisting}[
    basicstyle=\ttfamily\scriptsize,
    breaklines=true,
    breakatwhitespace=true,
    columns=flexible,
    keepspaces=true,
    xleftmargin=0pt,
    framexleftmargin=0pt,
    aboveskip=3pt,
    belowskip=3pt
]
{
  "Goal": "{{target_object_category}}",
  "Prior Probabilities between Room Type and 
   Goal Object": [
    {{prob_entries}}
  ],
  "Area Descriptions": [
    {{area_entries}}
  ]
}
\end{lstlisting}
    
    \end{tcolorbox}
    \label{tab:prompt-single}
\end{table}

In this work, we design structured prompts to facilitate zero-shot multi-floor object navigation. As illustrated in Tab.~\ref{tab:prompt-single} and Tab.~\ref{tab:prompt-multi}, these prompts include the following key components:
\begin{itemize}
    \item \textbf{Task Specification}: Defines the objective of selecting the optimal floor based on prior probabilistic data and environmental context.
    \item \textbf{Expected Output Format}: Specifies JSON format for machine-readable, consistent responses.
    \item \textbf{Example Input}: Provides a JSON example including the target object, prior probabilities for floors and room types, and floor descriptions.
    \item \textbf{Example Response}: Demonstrates the output format with a floor index and reasoning justification.
    \item \textbf{Template for Prompted Input}: Uses placeholders (\texttt{\{\{target\_object\_category\}\}}, \texttt{\{\{floor\_prob\_entries\}\}}, etc.) for flexible adaptation to different scenarios.
\end{itemize}
These components guide the model in leveraging probabilistic data and environmental context for zero-shot navigation, ensuring clarity, consistency, and effective decision-making in unseen environments.

\begin{table}[t]
    \centering
    \caption{An example of the instruction for multi-floor object navigation.}
    \renewcommand{\arraystretch}{1.2}
    \setlength{\tabcolsep}{8pt}
    \tcbset{
        colback=gray!10,
        colframe=gray!50,
        arc=2mm,
        boxrule=0.3mm,
        width=\linewidth,
        left=3pt,
        right=3pt,
        top=3pt,
        bottom=3pt,
        boxsep=0pt,
    }
    \begin{tcolorbox}
    \small
    \textbf{\texttt{Task: Select the optimal floor based on prior probabilistic data and environmental context.}}
    
    \vspace{0.15cm}
    
    \texttt{\textbf{Expected Answer Format: JSON}}
    
    \vspace{0.15cm}
    
    \texttt{\textbf{\textcolor{orange}{Example Input:}}}
    
\begin{lstlisting}[
    basicstyle=\ttfamily\scriptsize,
    breaklines=true,
    breakatwhitespace=true,
    columns=flexible,
    keepspaces=true,
    xleftmargin=0pt,
    xrightmargin=0pt,
    framexleftmargin=0pt,
    framexrightmargin=0pt,
    aboveskip=2pt,
    belowskip=2pt
]
{
  "Goal": "bed",
  "Prior Probabilities between Floor and Goal Object": [
    "Floor 1": 10.0%, "Floor 2": 10.0%, "Floor 3": 80.0%
  ],
  "Prior Probabilities between Room Type and Goal Object": [
    "Bedroom": 80.0%, "Living room": 15.0%, "Bathroom": 5.0%
  ],
  "Floor Descriptions": [
    "Floor 1": "Current floor. There are room types: hall, 
                living room, containing objects: tv, sofa",
    "Floor 2": "Other floor. There are room types: bathroom 
                containing objects: shower, towel. You do not 
                need to explore this floor again",
    "Floor 3": "Other floor. There are room types: unknown 
                rooms containing objects: unknown objects"
  ]
}
\end{lstlisting}

    \texttt{\textbf{\textcolor{orange}{Example Response:}}}
    
\begin{lstlisting}[
    basicstyle=\ttfamily\scriptsize,
    breaklines=true,
    breakatwhitespace=true,
    columns=flexible,
    keepspaces=true,
    xleftmargin=0pt,
    xrightmargin=0pt,
    framexleftmargin=0pt,
    framexrightmargin=0pt,
    aboveskip=2pt,
    belowskip=2pt
]
{
  "Index": "3",
  "Reason": "The bedroom is most likely to be on the Floor 3, 
             and the room types and object types on the Floor 1 
             and Floor 2 are not directly related to the target 
             object bed, especially it does not need to explore 
             Floor 2 again."
}
\end{lstlisting}

    \texttt{\textbf{\textcolor{brown}{Prompted Input:}}}
    
\begin{lstlisting}[
    basicstyle=\ttfamily\scriptsize,
    breaklines=true,
    breakatwhitespace=true,
    columns=flexible,
    keepspaces=true,
    xleftmargin=0pt,
    xrightmargin=0pt,
    framexleftmargin=0pt,
    framexrightmargin=0pt,
    aboveskip=2pt,
    belowskip=2pt
]
{
  "Goal": "{{target_object_category}}",
  "Prior Probabilities between Floor and Goal Object": [
    {{floor_prob_entries}}
  ],
  "Prior Probabilities between Room Type and Goal Object": [
    {{prob_entries}}
  ],
  "Floor Descriptions": [
    {{floor_entries}}
  ]
}
\end{lstlisting}
    
    \end{tcolorbox}
    \label{tab:prompt-multi}
\end{table}

\end{document}